\documentclass[sigconf,screen]{acmart}
\usepackage{amsmath}
\usepackage{graphicx}
\usepackage{booktabs}
\renewcommand\footnotetextcopyrightpermission[1]{}
\usepackage{multirow}
\usepackage{multicol}

\usepackage{epstopdf}
\AtBeginDocument{%
  }

\setcopyright{acmlicensed}
\copyrightyear{2026}
\acmYear{2026}
\acmDOI{XXXXXXX.XXXXXXX}

\begin{document}
\settopmatter{printfolios=true}

\title{SAMIC: A Lightweight Semantic-Aware Mamba for Efficient Perceptual Image Compression}

\author{Jiaqian Zhang*}
\affiliation{
  \institution{State Key Laboratory of Human-Machine Hybrid Augmented Intelligence, Institute of Artificial Intelligence and Robotics, Xi'an Jiaotong University}
  \city{Xi'an}
  \country{China}
}

\author{Hao Wei*}
\affiliation{
  \institution{State Key Laboratory of Human-Machine Hybrid Augmented Intelligence, Institute of Artificial Intelligence and Robotics, Xi'an Jiaotong University}
  \city{Xi'an}
  \country{China}
}

\author{Chenyang Ge}
\affiliation{
  \institution{State Key Laboratory of Human-Machine Hybrid Augmented Intelligence, Institute of Artificial Intelligence and Robotics, Xi'an Jiaotong University}
  \city{Xi'an}
  \country{China}
}

\author{Yanhui Zhou}
\affiliation{
  \institution{School of Information and Communication Engineering, Xi'an Jiaotong University}
  \city{Xi'an}
  \country{China}
}

\begin{abstract}
Perceptual image compression focuses on preserving high visual quality under low-bitrate constraints. Most existing approaches to perceptual compression leverage the strong generative capabilities of generative adversarial networks or diffusion models, at the cost of substantial model complexity. 
To this end, we present an efficient perceptual image compression method that exploits the long-range modeling capability and linear computational complexity of state space models, with a particular focus on Mamba. 
Unlike existing methods that rely on an inherently fixed scanning order and consequently impair semantic continuity and spatial correlation, we develop a semantic-aware Mamba block (SAMB) to enable scanning guided by dynamically clustered semantic features, thereby alleviating the strict causality constraints and long-range information decay inherent to Mamba.  
Inspired by singular value decomposition, we design an SVD-inspired redundancy reduction module (SVD-RRM) that performs a low-rank approximation on the latent features by introducing a learnable soft threshold, leading to channel-wise redundancy information reduction.
The proposed SAMB is integrated into both the encoder and decoder of the compression framework, whereas the SVD-RRM is incorporated only in the encoder.
Extensive experiments demonstrate that our method performs favorably against state-of-the-art approaches in terms of rate-distortion-perception tradeoff and model complexity. The source code and pretrained models will be available at https://github.com/Jasmine-aiq/SAMIC.
\end{abstract}



\keywords{Perceptual image compression, state space models, semantic-aware Mamba, singular value decomposition}


\maketitle

\section{Introduction}

In the current era of exponentially growing ultra-high-definition and high-resolution images, image compression technology has become a crucial component in scenarios constrained by storage capacity and transmission bandwidth. 

Traditional compression standards, such as JPEG \cite{JPEG_ACM1991}, BPG \cite{BPG_2015}, and VCC \cite{VCC_2021}, have played a significant role in advancing digitalization. However, due to their block‑based coding schemes, they are prone to introducing blocking artifacts and blurs at low bitrates. 

Early learned image compression (LIC) approaches \cite{End-to-end_ICLR2016, hyperprior_ICLR2018} primarily employ convolutional neural networks (CNNs) for nonlinear transforms. However, limited by their local receptive fields, CNNs struggle to capture long‑range dependencies, resulting in restricted rate‑distortion performance in high‑resolution scenarios. To address these issues, researchers have introduced Transformers \cite{Entroformer_ICLR2022, FAT_ICLR2024, MTCM_CVPR2023}, which leverage self‑attention mechanisms to model global context and further enhance compression efficiency. Nevertheless, their computational complexity grows quadratically with the number of pixels, leading to significant latency and computational overhead on high‑resolution images and severely hindering their practical deployment. While these methods achieve impressive performance on objective metrics, they often produce severely over-smoothed reconstructions at low bitrates. This occurs because the model tends to sacrifice high-frequency information in exchange for lower pixel-level errors, resulting in a significant loss of rich textural details.

\begin{figure}[t!]
    \centering
    \begin{minipage}{0.48\linewidth}
        \centering
        \includegraphics[width=\linewidth]{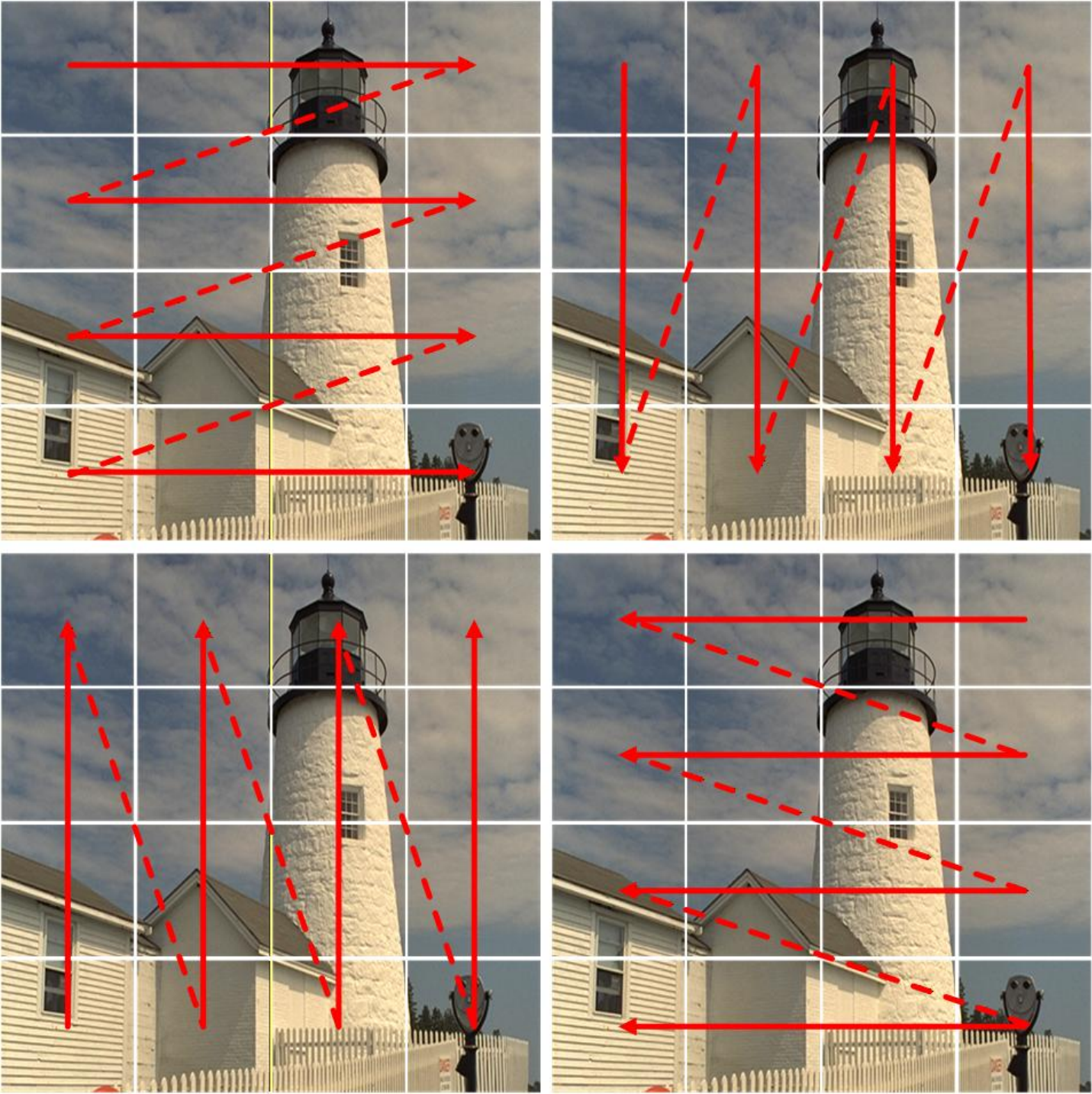}
        \centerline{\footnotesize (a) SS2D \cite{VMamba_NIPS2024}}
    \end{minipage}
    \hfill
    \begin{minipage}{0.48\linewidth}
        \centering
        \includegraphics[width=\linewidth]{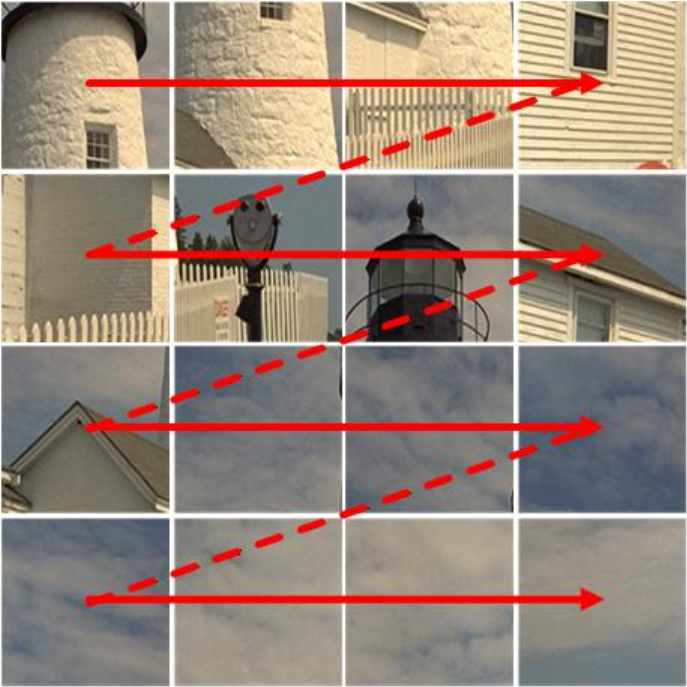}
        \centerline{\footnotesize (b) Cassic \cite{Cassic_ICCV2025}}
    \end{minipage}
    
    \vspace{0.5em}
    
    \begin{minipage}{0.48\linewidth}
        \centering
        \includegraphics[width=\linewidth]{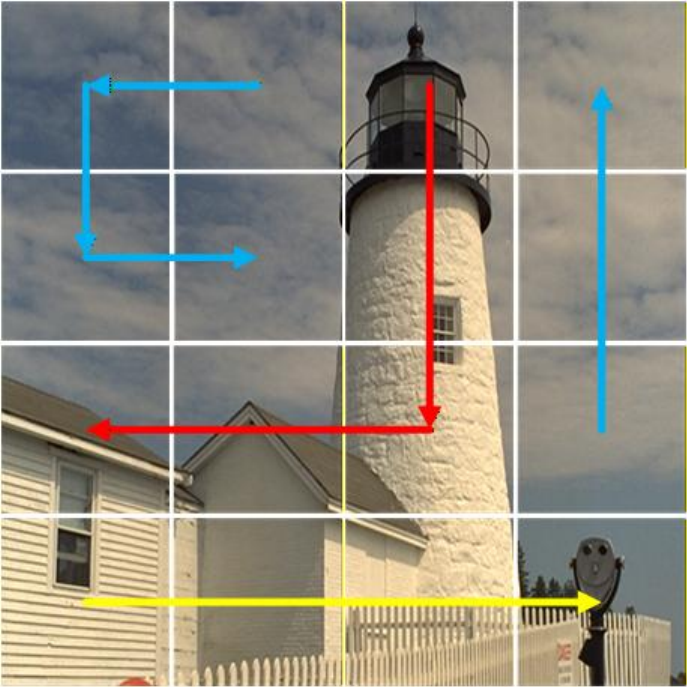}
        \centerline{\footnotesize (c) CMIC \cite{CMIC_arXiv2025}}
    \end{minipage}
    \hfill
    \begin{minipage}{0.48\linewidth}
        \centering
        \includegraphics[width=\linewidth]{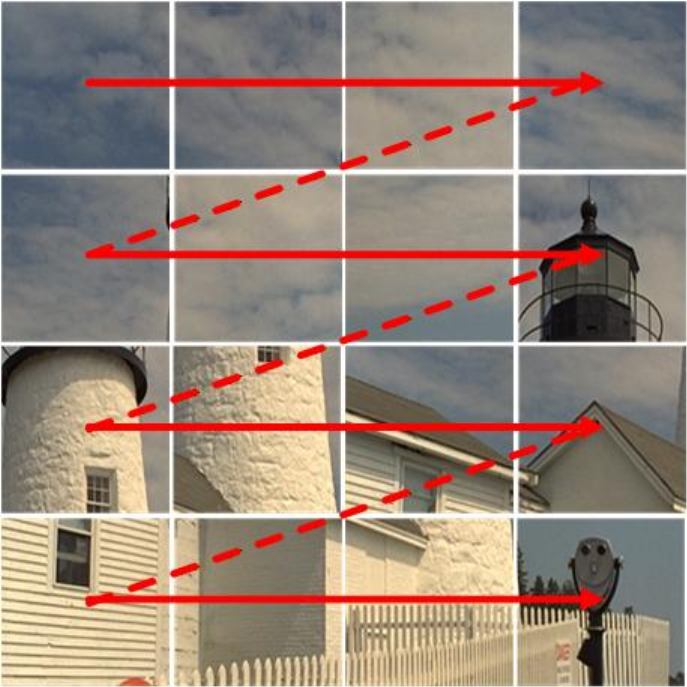}
        \centerline{\footnotesize (d) SAMIC}
    \end{minipage}
    \caption{\textbf{Comparison of scanning strategies in Mamba.} (a) The standard 2D \cite{VMamba_NIPS2024} selective scan (SS2D) adopts a fixed, inherent raster-scanning order. This rigid trajectory strictly follows spatial grids but severely disrupts the semantic continuity of distinct objects (e.g., the lighthouse or the sky). (b) Cassic \cite{Cassic_ICCV2025} employs weighted activation map and bit allocation map to guide content-adaptive selective scan at encoder and decoder, respectively. (c) CMIC \cite{CMIC_arXiv2025} adopts a content-adaptive strategy by reorganizing tokens based on low-level feature similarity. (d) Our proposed semantic-aware selective scanning dynamically constructs a content-adaptive scanning trajectory. By clustering semantically similar patches, it ensures that regions sharing identical semantic information are processed contiguously.}
    \label{fig:1}
\end{figure}

In recent years, perceptual image compression has emerged as an important research direction. Inspired by the success of generative adversarial networks (GANs) \cite{GAN_CACM2020}, some methods \cite{HiFiC_NIPS2020, GANextreme_ICCV2019, PoELIC_CVPR2022} treat the decoder as a generator and introduce a discriminator to enhance the visual realism of reconstructed images. These approaches have achieved noticeable improvements in subjective visual quality. However, in the absence of suitable priors, they often rely on large‑scale decoder models, which substantially increase parameter counts and computational complexity, limiting their practicality in storage and power‑constrained devices. To further boost perceptual quality, several works have incorporated explicit semantic priors, such as semantic segmentation maps \cite{DSSLIC_ICASSP2019}, or textual descriptions \cite{MMDN_AAAI2023, TACO_ICML2024} into the encoder and decoder. 
However, methods relying on explicit semantic priors usually require additional semantic extraction modules or auxiliary models, introducing extra encoding and inference overhead and increasing overall system complexity. 

As demonstrated in recent studies, state space models—most notably Mamba—are capable of capturing global dependencies with linear complexity \cite{S4_ICLR2022, Mamba_COLM2024, VisionMamba_ICML2024, VMamba_NIPS2024}. This capability has motivated a growing number of researchers to adopt them for image compression. 
For example, MambaIC \cite{MambaIC_CVPR2025} used a 2D selective scan borrowed from \cite{VMamba_NIPS2024} to capture long-range dependency (see Fig.~\ref{fig:1} (a)). Nevertheless, this predefined fixed scanning strategy cannot accommodate variations in image content, which results in inefficient context modeling and consequently wastes model capacity on irrelevant dependencies.
%
To mitigate this shortcoming, Cassic \cite{Cassic_ICCV2025} and CMIC \cite{CMIC_arXiv2025} approaches begin to perform adaptive scanning according to the content of images. Specifically, Cassic method unfolds 2D features into a 1D sequence based on the descending order of the bit allocation map before scanning (Fig.~\ref{fig:1} (b)), while CMIC employs K-means clustering to group tokens based on feature similarity (Fig.~\ref{fig:1} (c)).
%

By contrast, we propose a lightweight perceptual image compression framework that achieves semantic-aware scanning through soft clustering and Gumbel-Softmax without relying on external clustering algorithms, and employs explicitly embedding 2D positional encodings to preserve spatial continuity. We introduce a semantic-aware scanning strategy within the Mamba blocks of the encoder, decoder, and context entropy model, which dynamically reorganizes the input sequence to preserve semantic proximity and spatial continuity during state-space modeling. Crucially, this improvement requires few parameters during training and inference. Furthermore, we introduce an SVD-inspired redundancy reduction module after the main encoder to explicitly remove channel redundancy via low-rank approximation, allowing for a more compact model representation without sacrificing performance. Extensive experiments demonstrate that the proposed framework achieves competitive rate–distortion–perception performance with only 25M parameters, outperforming most existing perceptual image compression methods in both objective metrics (PSNR, MS-SSIM) and perceptual quality (LPIPS), particularly in low-bitrate scenarios. The main contributions are summarized as:

\begin{itemize}
\item We propose a lightweight perceptual image compression framework based on the Mamba that achieves a favorable trade-off between compression efficiency, perceptual quality, and computational complexity.
\item We introduce a semantic-aware Mamba that ensures semantic-spatial continuity, expands the effective receptive field, and maintains linear complexity without extra clustering.
\item An SVD-inspired redundancy reduction module is developed and incorporated into the encoder. Within this module, a learnable soft-thresholding operation is applied to perform singular value decomposition on the latent representation, achieving low-rank approximation and explicitly removing channel redundancy. 
\end{itemize}
\begin{figure*}[t!]
    \centering
    \includegraphics[width=0.95\textwidth,height=9cm,keepaspectratio]{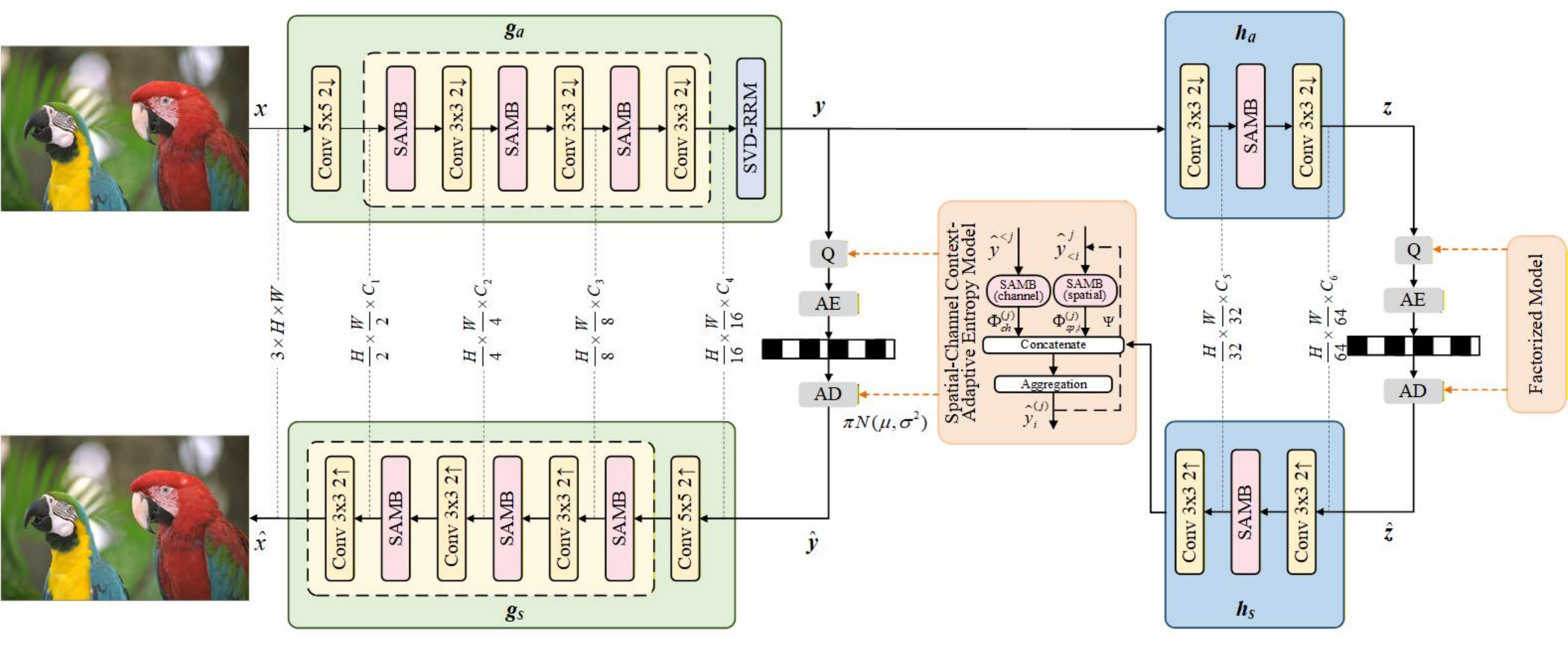}
    \caption{\textbf{Architecture of the proposed SAMIC framework}. Specifically, we develop an effective semantic-aware Mamba block (SAMB) as the core module of the encoder and decoder. Moreover, an SVD-inspired channel-wise redundancy reduction module (SVD-RRM) is incorporated into the encoder for more compact feature representation.}
    \label{fig:2}
\end{figure*}

\section{Related Work}
\subsection{Perceptual learned compression methods}
Deep learning-based image compression has achieved remarkable progress in terms of rate-distortion tradeoff. 
Ballé et al. \cite{End-to-end_ICLR2016, hyperprior_ICLR2018} pioneered an end‑to‑end learnable compression framework, jointly training the encoder, quantizer, and decoder through differentiable rate‑distortion optimization. Subsequently, Minnen et al. \cite{minnen2018joint} further introduced a context‑adaptive entropy model, significantly improving the accuracy of entropy coding. Moreover, Transformer-based compression approaches have been proposed to overcome the limitations of the local receptive field inherent in previous CNN-based methods \cite{WAIC_CVPR2022, Entroformer_ICLR2022, FAT_ICLR2024, STF_CVPR2022}. In \cite{MTCM_CVPR2023}, Liu et al. developed a Transformer-CNN hybrid architecture for compression. However, these methods typically optimize for distortion metrics such as PSNR, which cannot fully capture the subjective perceptual characteristics of the human visual system.

Perceptual image compression methods use perceptual losses to enhance the perceptual quality of reconstructions at low bitrates \cite{GANextreme_ICCV2019, HiFiC_NIPS2020, MS-ILLM_PMLR2023, ICISP_NN2025}. HiFiC \cite{HiFiC_NIPS2020} emphasizes the detailed design of its generator, discriminator, and training strategies. Muckley et al. \cite{MS-ILLM_PMLR2023} used the same generator as the HiFiC and proposed a non-binary discriminator conditioned on quantized local image representations obtained through a VQVAE autoencoder \cite{VQVAE_NIPS2017}. Wei et al. further improved the discriminator by introducing the implicit priors \cite{ICISP_NN2025}.


Recently, diffusion models have also been employed for perceptual image compression~\cite{CGDM_ACMMM2024, RDEIC_TCSVT2025}. CDC \cite{CDC_NIPS2023} method used a conditional diffusion model as its decoder, conditioned on the compressed latent features. Li et al. \cite{DiffEIC_TCSVT2024} proposed a lightweight control module to improve reconstruction quality under extremely low bitrates, which connects the compressed latents to a frozen stable diffusion model. To further improve efficiency, OSDiff \cite{OSDiff_VCIP2025} achieved one-step diffusion compression by designing a discriminator based on compact feature representations instead of raw pixels.
%
Nevertheless, diffusion‑based methods generally depend on iterative sampling and incur significant model complexity, leading to considerable computational and memory overhead. This renders them difficult to deploy on resource‑constrained end‑user devices.

\subsection{State Space Models}
%
The state space model (SSM), as an emerging paradigm for sequence modeling, has recently demonstrated significant potential in efficiently modeling long‑range dependencies. 
Mamba extends the traditional SSM architecture by incorporating an input‑dependent selection mechanism, enabling stronger content‑adaptive modeling while maintaining linear time complexity. 
Consequently, this trend has inspired a growing number of researchers to investigate the potential of Mamba in image compression.
MambaVC \cite{MambaVC_arXiv2024} designed a visual state space block inherited from vision Mamba \cite{VMamba_NIPS2024} as the core module of the encoder and decoder. Similarly, Zeng et al. \cite{MambaIC_CVPR2025} also adopted the 2D selective scanning from \cite{VMamba_NIPS2024} for non-local modeling. However, these methods cannot preserve spatial continuity and local structural consistency—two properties essential for image compression. 

To mitigate the limitations noted above, Wei et al. \cite{ICISP_NN2025} introduced local convolution to help capture local dependency. Chen et al. \cite{CMIC_arXiv2025} introduced a prompt dictionary to embed global priors, thereby addressing the strict causality constraint without the need for four-directional scanning. Cassic method~\cite{Cassic_ICCV2025} enabled content-adaptive selective scan according to weighted activation maps and bit allocation maps. In contrast, we develop a semantic-aware Mamba block that performs semantic-guided selective scanning based on a semantic clustering strategy.

\section{Method}

\subsection{Overview}
\label{overview}
The overall compression pipeline is shown in Fig.~\ref{fig:2}. Given an input image $x$, it is encoded into a latent representation $\text{y}$ through the main encoder $\text{g}_a$: 
$\text{y}=\text{g}_a(x)$. Then, $\text{y}$ is processed by the hyper‑prior encoder $h_a$ to obtain the hyper‑latent $z$: $z=h_a(\text{y})$. 
The quantized hyper‑latent is denoted as $\hat{z}=Q(z)$, where $Q(\cdot)$ represents the quantization operation. Then, $\hat{z}$ is fed into the hyper‑prior decoder $h_s$ to produce the mean $\mu$ and variance $\sigma^2$ for the latent $\text{y}$: $\mu,\sigma=h_s(\hat{z})$. The quantized latent is produced by $\hat{\text{y}}=Q(\text{y})$, which is modeled as a conditional Gaussian distribution: $p_{\hat{\text{y}}|\hat{z}}(\hat{\text{y}}|\hat{z})=\mathcal{N}(\hat{\text{y}}|\mu,\sigma^2 I)$ by a spatial‑channel context‑adaptive entropy model. Note that the entropy model is adopted from \cite{MambaIC_CVPR2025}, but with their Mamba block replaced by our semantic-aware Mamba block. Finally, the reconstructed image $\hat{x}$ is obtained by passing $\hat{\text{y}}$ through the main decoder $\text{g}_s$: $\hat{x}=\text{g}_s(\hat{\text{y}})$.

To be specific, we develop a semantic-aware Mamba block that performs a semantic‑aware selective scanning mechanism for semantic continuity enhancement, formulated as the core component of the encoder and decoder.
 Furthermore, we design a learnable channel-wise redundancy reduction module based on singular value decomposition. This module performs low‑rank approximation on the latent representation $\text{y}$, effectively reducing channel‑wise redundancy and achieving higher compression efficiency.
In the following, we elaborate on the aforementioned delicate blocks.

\begin{figure*}
    \centering   \includegraphics[width=\textwidth,height=4.3cm]{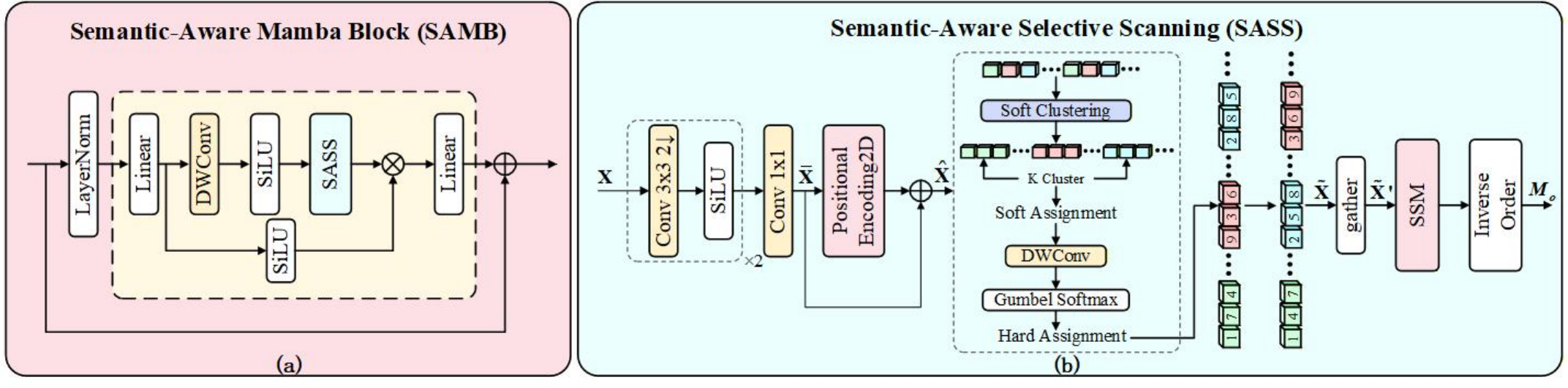}      
    \caption{Detailed architecture. (a) Semantic-aware Mamba block (SAMB) adopts the proposed SASS for global modeling. (b) Semantic-aware selective scanning (SASS) generates pixel-wise hard assignments using soft clustering and Gumbel Softmax. Based on these assignments, the features are reordered to ensure semantic continuity, followed by an inverse order operation to restore the spatial layout.}
    \label{fig:3}
\end{figure*}

\subsection{Semantic-Aware Mamba}
\label{Semantic-Aware Mamba}
To overcome the semantic discontinuity issue caused by fixed scanning orders in traditional vision Mamba, we propose a semantic-aware Mamba block (SAMB) that involves a novel scanning strategy, which adaptively constructs the scanning order based on the semantic content of the image. This strategy simultaneously considers both feature‑space proximity and Euclidean spatial proximity, effectively alleviating the strict causality and long‑range decay of vanilla Mamba while maintaining linear complexity.
\subsubsection{Semantic Feature Extraction}
The detailed architecture of the proposed SAMB is illustrated in Fig.~\ref{fig:3} (a). Different from VMamba~\cite{VMamba_NIPS2024}, we replace the 2D selective scan module with our semantic-aware selective scanning approach (SASS). Specifically, SASS introduces a learnable semantic feature extraction module to dynamically generate a content‑adaptive scanning order. As shown in Fig.~\ref{fig:3} (b), given an input feature map $\mathbf{X}$, a lightweight convolutional network is first employed to extract pixel‑wise semantic features, followed by a $1\times1$ convolution that projects the features into a low‑dimensional semantic space:
\begin{align}
\mathbf{\bar{X}}=\text{Conv}_{1\times1}(\sigma(\text{Conv}_{3\times3}(\sigma(\text{Conv}_{3\times3}(\mathbf{X}))))), \label{eq:1}
\end{align}
where $\text{Conv}_{s\times s}$ denotes the convolution with a kernel size of $s$, $\sigma$ is SiLU activation, and $\bar{\textbf{X}}$ is output feature.

To preserve spatial structure, we introduce 2D sinusoidal positional encoding, which helps the clustering process perceive the relative spatial relationships among pixels:
\begin{equation}
    \mathbf{\hat{X}}=\mathbf{\bar{X}}+\sin(\pi v) \oplus \cos(\pi u), \label{eq:3}
\end{equation}
where $u$ and $v$ represent positional information, respectively.

\subsubsection{Semantic Clustering}
Before performing scanning, we need to group pixels into K semantic clusters. Specifically, we first define a set of learnable cluster centers $\{c_k\}_{k=1}^{K} \subset \mathbb{R}^C_{sem}$. 
For each position of feature $\mathbf{\hat{X}}$, we compute its cosine similarity with all cluster centers and obtain a soft assignment probability distribution via a softmax function as:
\begin{equation}
    s_{k,i} = \frac{\mathbf{\hat{X}}_{i}^\top c_k}{\|\mathbf{\hat{X}}_{i}\| \cdot \|c_k\| \cdot \tau}, \quad
    p_{k,i} = \frac{\exp(s_{k,i})}{\sum_{k'=1}^{K} \exp(s_{k',i})},
\end{equation}
where $\tau$ is the temperature coefficient, and $p_{k,i}$ denotes the soft assignment probability of the $i$-th pixel $\mathbf{\hat{X}}_{i}$ of feature $\mathbf{\hat{X}}$ belonging to the $k$-th cluster. 
Then, an approximate one‑hot hard assignment $q$ is derived using Gumbel‑Softmax, $q=\text{Gumbel-Softmax}\bigl(p\bigr)$. Based on the hard assignment $q\in\{0,1\}$, a scanning order is constructed independently for each image.

Specifically, for each cluster $k$, we compute its average spatial position as the sorting key:
\begin{equation}
\bar{u}_k = \frac{\sum_{i=1}^{H\times W} q_{k,i} \cdot u_i}{\sum_{i=1}^{H\times W} q_{k,i}}, \quad
\bar{v}_k = \frac{\sum_{i=1}^{H\times W} q_{k,i} \cdot v_i}{\sum_{i=1}^{H\times W} q_{k,i}},
\end{equation}
where $ (u_i, v_i) $ is the spatial coordinate of pixel $i$ in feature $\mathbf{\hat{X}}$ with a spatial resolution of $H\times W$. Then, the $K$ clusters are sorted in ascending order according to their sorting key. Within each cluster, pixels are arranged in their original raster order to generate the global scanning sequence, which ensures that semantically similar regions appear contiguously while preserving the overall spatial layout of the image.

\subsubsection{Semantic-guided Selective Scanning}
Based on the hard assignment $q$ obtained from Gumbel-Softmax, we generate a permutation index $\pi$ that sorts pixels according to their cluster identity. The input feature $\mathbf{\tilde{X}}$ are then rearranged into a 1D sequence $\mathbf{\tilde{X}'}$ using the Gather operation:
\begin{equation}
\mathbf{\tilde{X}'} = \text{Gather}(\mathbf{\tilde{X}}, \pi),
\end{equation}
Gather operation along the spatial dimension, where  
$ \tilde{X}[i]= \hat{X}[\pi[i]]$, denotes that the $i$-th element of the sequence corresponds to the $\pi[i]$-th pixel of the original spatial grid. Inside the core SASS, we employ a single scanning direction to minimize computational overhead. The reordered sequence is sent to the standard SSM to model global dependencies, and restored to its original 2D spatial grid using Gather operation as:
\begin{equation}
\mathbf{M}_{o} = \text{Gather}(\text{SSM}(\mathbf{\tilde{X}'}), \pi^{-1}),
\end{equation}
where $\pi^{-1}$ is the inverse version of $\pi$.

\begin{figure}
    \centering   \includegraphics[width=0.45\textwidth,height=4.5cm]{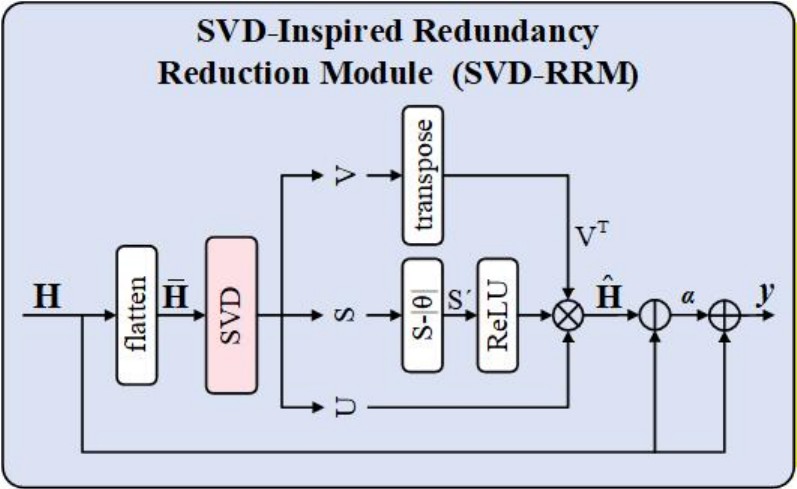}   
    \caption{
Detailed architecture of the proposed. SVD-inspired redundancy reduction Module (SVD-RRM). It employs a learnable soft-thresholding operation to remove redundancy and reconstructs the feature via a residual connection scaled by a learnable factor.}
    \label{fig:4}
\end{figure}

\begin{figure*}
    \centering       \includegraphics[width=1.0\textwidth,height=5cm,keepaspectratio]{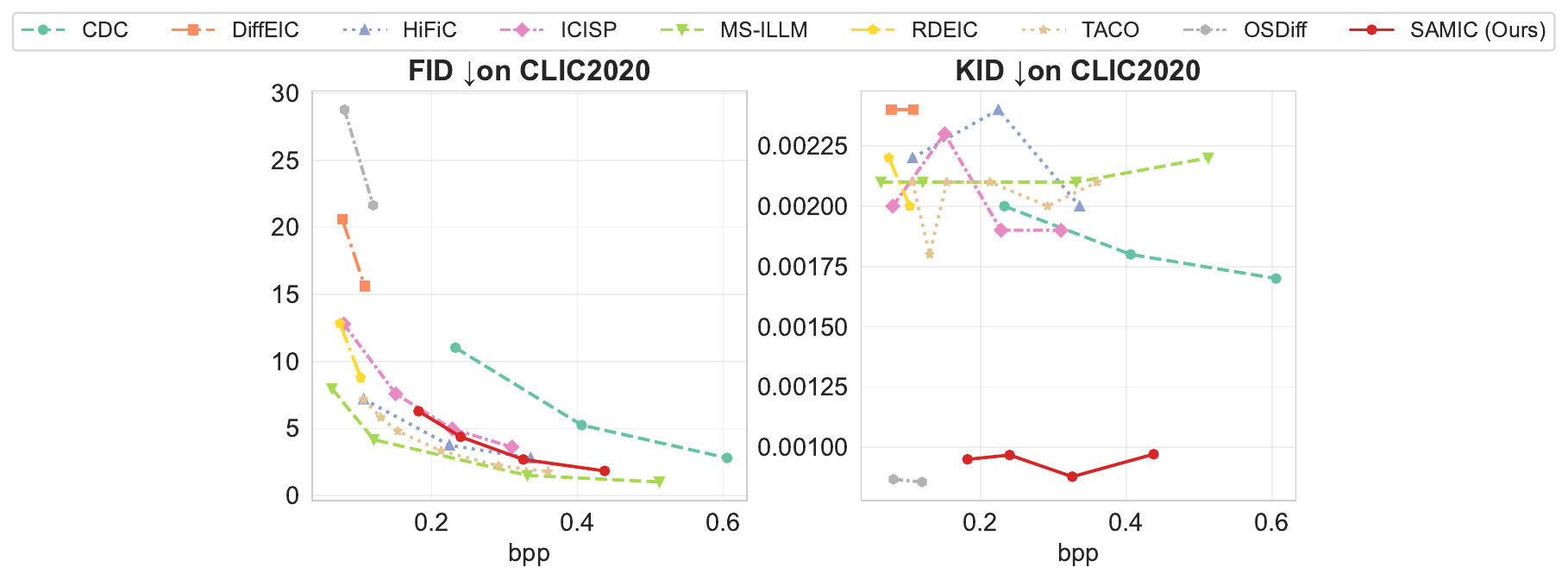} \includegraphics[width=1.0\textwidth,height=9cm,keepaspectratio]{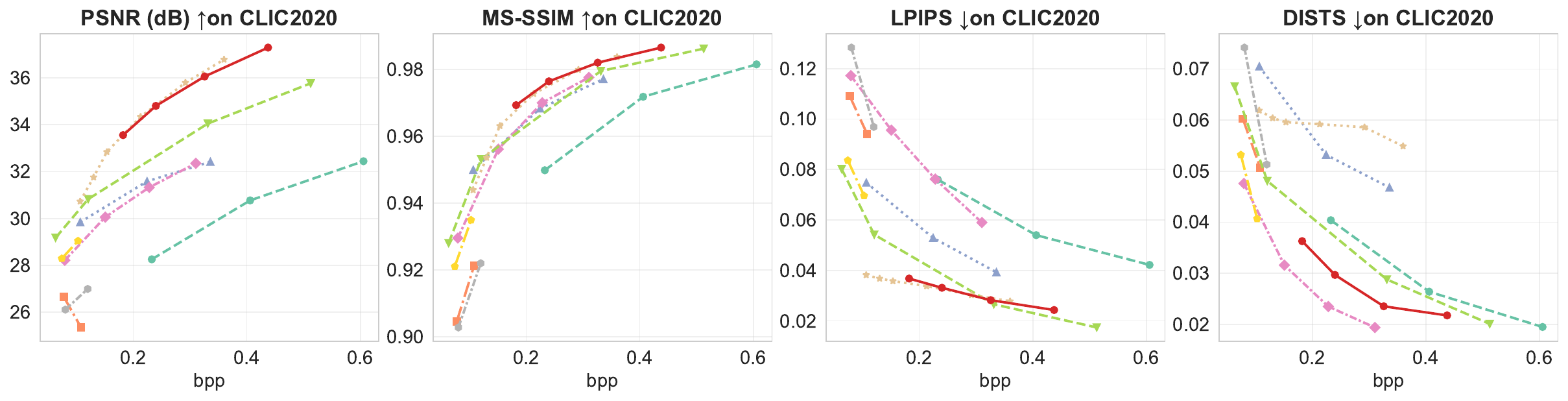}    \includegraphics[width=1.0\textwidth,height=9cm,keepaspectratio]{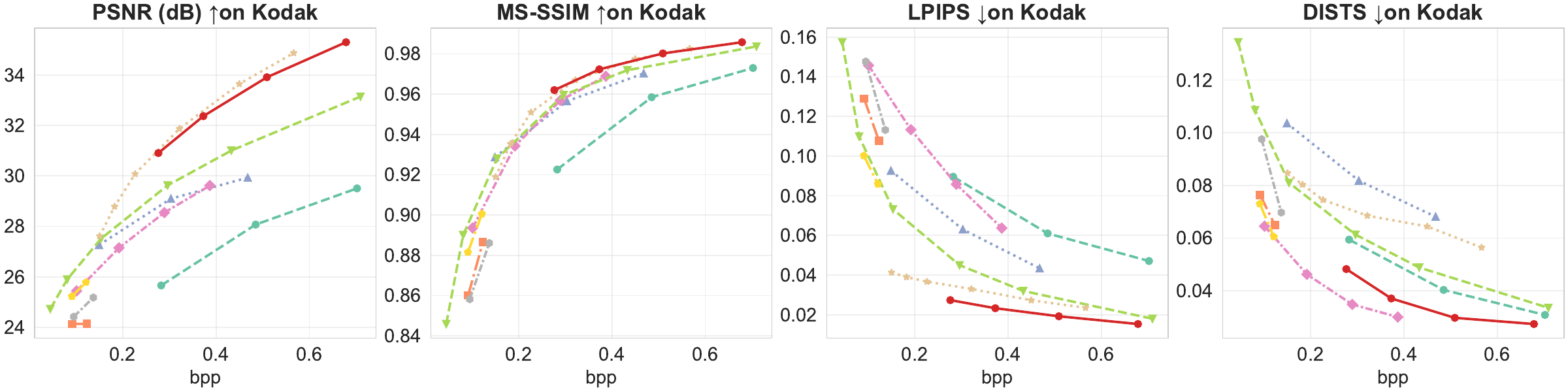}
    \caption{Quantitative rate-distortion-perception performance comparisons on the Kodak and CLIC2020 benchmark datasets. Higher PSNR/MS-SSIM and lower LPIPS/DISTS/KID/FID values indicate better performance. The proposed SAMIC achieves a better balance rate-distortion-perception tradeoff, while competing methods tend to favor one metric at the expense of another.} 
    \label{fig:5}
\end{figure*}

\subsection{SVD-Inspired Redundancy Reduction}
\label{SVD-Inspired Redundancy Reduction Module}
In image compression, the latent representation $\text{y}$ output by the encoder often exhibits significant channel redundancy, resulting in many singular values that are close to zero or contribute minimally. This redundancy not only wastes bit resources but may also introduce noise that interferes with subsequent entropy coding and decoding processes.

To explicitly remove channel‑wise redundancy, we propose a lightweight SVD-inspired redundancy reduction module that is formulated as a lightweight and learnable SVD‑based low‑rank approximation. This block performs singular value decomposition on the latent feature matrix, retains the principal singular components via soft‑threshold truncation, and incorporates a learnable residual scaling factor, enabling adaptive compression of channel redundancy.

To be specific, given the latent feature $\mathbf{H} \in \mathbb{R}^{C \times H \times W}$ with the spatial dimension of $H\times W$ and channel dimension of $C$, we first flatten it as $\mathbf{\bar{H}}\in \mathbb{R}^{C\times N}$, where $N=H \times W$ denotes the number of pixels. Subsequently, we compute the SVD of $\mathbf{\bar{H}}$ as:
\begin{equation}
     \textbf{S}, \textbf{U}, \textbf{V}=\text{SVD}(\mathbf{\bar{H}}), 
\end{equation}
where $\textbf{U}$ and $\textbf{V}$ denote the left and right orthogonal singular vector matrices, respectively. $\textbf{S}=[s_1, s_2, \dots, s_{min(C,N)}]$ represents the singular values in descending order. 

Then a learnable soft‑thresholding operation is applied to $\textbf{S}$, allowing the model to adaptively decide how many singular components to retain during training. 
\begin{equation}
    \textbf{S}'= \max (s_i-\theta,0),  \quad \text{i}\in[1, min(C,N)],  \\
\end{equation}
where $\theta$ is a learnable parameter and $\textbf{S}'$ denotes the truncated singular value vector.
 
Finally, the compact feature $\mathbf{\hat{H}}$ is reconstructed via an element-wise multiplication operation as:
\begin{equation}
     \mathbf{\hat{H}} = \mathbf{U} \cdot \textbf{S}' \cdot \mathbf{V}^\top,
\end{equation}
To avoid excessive information loss due to over‑compression, we introduce a learnable scaling factor $\alpha$ that controls the strength of the low‑rank correction as follows:
\begin{equation}
    \text{y} = \mathbf{H} + \alpha \cdot(\mathbf{\hat{H}}-\mathbf{H}).
\end{equation}

\subsection{Context-adaptive Entropy Model}

Our entropy model builds upon the spatial-channel context modeling architecture proposed in \cite{MambaIC_CVPR2025}. We replace the VSS block with our proposed SAMB to better capture semantic and spatial continuity while mitigating redundancy.

Given latent features $\hat{\mathbf{y}}$, we partition the channels into $J$ chunks, indexed by $j$. For the decoding of the symbol at the $i$-th location of the $j$-th chunk, the channel context $\Phi_{ch}^{(j)}$ and spatial context $\Phi_{sp,i}^{(j)}$ are computed from previously decoded chunks $\hat{\mathbf{y}}^{<j}$ and previously decoded spatial symbols $\hat{\mathbf{y}}_{<i}^{j}$ in the current chunk, respectively. We utilize our SAMB module to capture these global receptive fields:
\begin{equation}
  \Phi_{ch}^{(j)} = \text{SAMB}(\hat{\mathbf{y}}^{<{j}})), \quad
  \Phi_{sp, i}^{(j)} = \text{SAMB}(\hat{\mathbf{y}}_{<i}^{j}).  
\end{equation}

The hyperprior context $\Psi$ is obtained by applying a nonlinear transform function $h_s$ to the hyperprior latent $\hat{\mathbf{z}}$: $\Psi = h_s(\hat{\text{z}})$. While the SAMB effectively extracts global dependencies and maintains semantic continuity, we additionally employ window-based local attention (WLA) for auto-regressive modeling to enhance fine-grained local relation descriptions. All three contexts are combined through a parameter aggregation network, followed by the WLA module to predict the final Gaussian entropy parameters $(\boldsymbol{\mu}_i^{j}, \boldsymbol{\sigma}_i^{j})$:
\begin{equation}
  (\boldsymbol{\mu}_i^{j}, \boldsymbol{\sigma}_i^{j}) = \mathcal{N}(\text{Concat}(\Phi_{ch}^{(j)}, \Phi_{sp, i}^{(j)}, \Psi)),  
\end{equation}
where $\text{Concat}$ denotes the channel-wise concatenation, $\mathcal{N}$ represents a network that consists of a $5 \times 5$ checkerboard masked convolution for spatial context prediction followed by a WLA module.

\subsection{Training Objective Function}
The overall training objective is formulated as a weighted combination of the rate-distortion-perceptual trade-off:
\begin{equation}
\label{loss}
    \mathcal{L} = \lambda \cdot R + \mathcal{D} + \beta \cdot \mathcal{P},
\end{equation}
where $R$ is the estimated bitrate, $\mathcal{D}$ is the distortion term, $\mathcal{P}$ is the perceptual loss, $\lambda$ and $\beta$ are hyper parameters that control the rate-distortion-perceptual trade-off.

The rate term is computed from the sum of the negative log-likelihoods of the learned entropy models: 
\begin{equation}
    R = -\frac{1}{\log 2 \cdot N_\text{pix}} \sum \log_2 p(\hat{\text{y}} | \psi),
\end{equation}
where $N_{pix}$ is the total number of pixels in the input image, and the summation is over all likelihoods provided by the entropy coding modules.

To enhance visual perceptual quality at low bitrates, a perceptual loss based on the LPIPS metric is incorporated:
\begin{equation}
    \mathcal{P} = \mathit{LPIPS}(\hat{x}, x),
\end{equation}
where LPIPS is computed using a pre-trained AlexNet backbone.

\section{Experiment}

\subsection{Setups}

\subsubsection{Datasets}
Our method was trained on the LSDIR \cite{LSDIR_CVPR2023} dataset, which contains 84,991 high-quality images, and then evaluated on the Kodak \cite{Kodak_1999} and CLIC2020 \cite{CLIC_2020} datasets. The Kodak dataset consists of 24 natural images with a resolution of 768$\times$512, while CLIC2020 comprises 428 images with a higher resolution. 

\subsubsection{Evaluation Metrics}
We measure the bitrate in bits per pixel (bpp) and use PSNR and MS-SSIM \cite{MS-SSIM_SSP2003} as reference metrics for distortion assessment. Additionally, we employ perceptual metrics LPIPS \cite{LPIPS_CVPR2018} and DISTS \cite{DISTS_TPAMI2020} to evaluate the visual quality of the reconstructions. FID \cite{heusel2017gans} and KID \cite{binkowski2018demystifying} metrics are used to evaluate the realism of the reconstructions.
Since the Kodak dataset is too small for reliable FID and KID value calculations, we do not report them for the Kodak dataset.

\subsection{Implementation Details} 
Our method was implemented using PyTorch framework and trained on a NVIDIA GeForce RTX 4090 GPU. The model was trained for 150 epochs. We optimized the network using the Adam optimizer with an initial learning rate of $1 \times 10^{-4}$, the learning rate was decayed by a factor of 0.1 at epochs 140 and 145. The training images were randomly cropped to $256\times256$ patches, and the batch size was set to 8. The number of features for $\text{y}$ and $z$ was set to 192 and 64, respectively. Note that the number of clusters K is set to 16. In Fig.~\ref{fig:2}, the intermediate features $\left\{C_1,C_2,C_3,C_4,C_5, C_6\right\}$ of our compression network are set as $\left\{128, 128, 128, 192, 128, 192\right\}$. The hyper-parameter $\beta$ in Eq. (\ref{loss}) was set to 3.5. To obtain compression models with different bitrates, the parameter $\lambda$ in Eq. (\ref{loss}) is selected from $\left\{ 0.0035, 0.0067, 0.013, 0.025 \right\}$.

\begin{figure*}[t]
    \centering
    \begin{minipage}[b]{\linewidth}
        
        
        \begin{minipage}[b]{0.99\linewidth}
            \centering
            \begin{minipage}[b]{0.13\linewidth}
                \includegraphics[width=\linewidth]{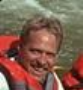}
                \centerline{(a1) GT (24)}
            \end{minipage}
            \hfill
            \begin{minipage}[b]{0.13\linewidth}
                \includegraphics[width=\linewidth]{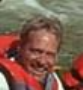}
                \centerline{(b1) SAMIC \textbf{(0.351)}}
            \end{minipage}
            \hfill
            \begin{minipage}[b]{0.13\linewidth}
                \includegraphics[width=\linewidth]{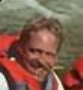}
                \centerline{(c1) TACO (0.422)}
            \end{minipage}
            \hfill
            \begin{minipage}[b]{0.13\linewidth}
                \includegraphics[width=\linewidth]{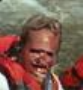}
                \centerline{(d1) CDC (0.363)} 
            \end{minipage}
            \hfill
            \begin{minipage}[b]{0.13\linewidth}
                \includegraphics[width=\linewidth]{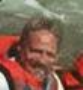}
                \centerline{(e1) HiFiC (0.376)}
            \end{minipage}
            \hfill
            \begin{minipage}[b]{0.13\linewidth}
                \includegraphics[width=\linewidth]{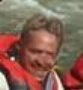}
                \centerline{(f1) ICISP (0.381)}
            \end{minipage}
            \hfill
            \begin{minipage}[b]{0.13\linewidth}
                \includegraphics[width=\linewidth]{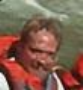}
                \centerline{(g1) MS-ILLM (0.353)}
            \end{minipage}
        \end{minipage} 
        
        \vspace{4pt}
        
        \begin{minipage}[b]{0.99\linewidth}
            \centering
            \begin{minipage}[b]{0.13\linewidth}
                \includegraphics[width=\linewidth]{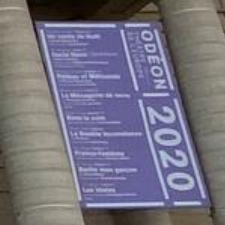}
                \centerline{(a2) GT (24)}
            \end{minipage}
            \hfill
            \begin{minipage}[b]{0.13\linewidth}
                \includegraphics[width=\linewidth]{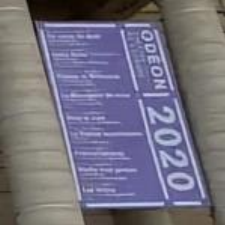}
                \centerline{(b2) SAMIC \textbf{(0.232)}}
            \end{minipage}
            \hfill
            \begin{minipage}[b]{0.13\linewidth}
                \includegraphics[width=\linewidth]{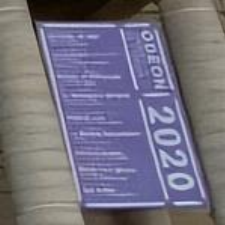}
                \centerline{(c2) TACO (0.266)}
            \end{minipage}
            \hfill
            \begin{minipage}[b]{0.13\linewidth}
                \includegraphics[width=\linewidth]{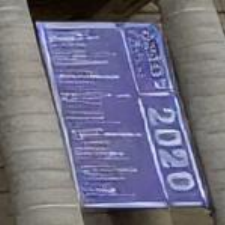}
                \centerline{(d2) CDC (0.288)} 
            \end{minipage}
            \hfill
            \begin{minipage}[b]{0.13\linewidth}
                \includegraphics[width=\linewidth]{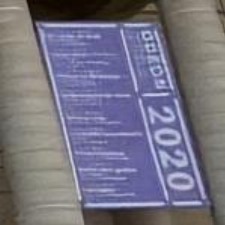}
                \centerline{(e2) HiFiC (0.254)}
            \end{minipage}
            \hfill
            \begin{minipage}[b]{0.13\linewidth}
                \includegraphics[width=\linewidth]{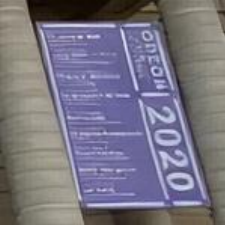}
                \centerline{(f2) ICISP (0.273)}
            \end{minipage}
            \hfill
            \begin{minipage}[b]{0.13\linewidth}
                \includegraphics[width=\linewidth]{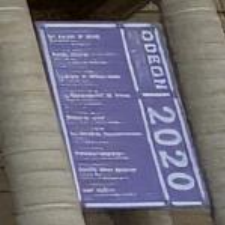}
                \centerline{(g2) MS-ILLM (0.261)}
            \end{minipage}
        \end{minipage}
    \end{minipage}
    \caption{\textbf{Visual comparisons on the Kodak and CLIC2020 datasets.} The value inside the parentheses represents bpp. SAMIC recovers high-frequency details and structural textures more faithfully than competing methods at lower bitrate.}
    \label{fig:6}
\end{figure*}

\subsection{Main Results}
We compared the proposed SAMIC method with state-of-the-art image compression approaches, including GAN-based methods (HiFiC \cite{HiFiC_NIPS2020}, MS-ILLM \cite{MS-ILLM_PMLR2023}, and ICISP \cite{ICISP_NN2025}), diffusion-based approaches (CDC \cite{CDC_NIPS2023}, DiffEIC \cite{DiffEIC_TCSVT2024}, RDEIC \cite{RDEIC_TCSVT2025}, and OSDiff \cite{OSDiff_VCIP2025}), and text-guided method TACO \cite{TACO_ICML2024}.

As shown in Fig.~\ref{fig:5}, the proposed SAMIC achieves competitive performance, ranking second only to the TACO method in terms of PSNR and MS-SSIM. However, it surpasses TACO on both LPIPS and DISTS metrics, indicating superior perceptual quality in its reconstructions. It is worth noting that while the ICISP method achieves the lowest DISTS value on the benchmark datasets, this comes at the cost of pixel fidelity, as evidenced by its lower PSNR compared to our SAMIC.

The visual comparisons are shown in Fig.~\ref{fig:6}. Our SAMIC method yields sharper reconstructions at lower bitrate compared to other approaches, allowing facial features and textual content to be clearly identified.

\subsection{Complexity comparisons}
As shown in Table~\ref{tab:1}, SAMIC demonstrates significant advantages in terms of model complexity, where our method has fewer network parameters, lower FLOPs, and faster inference speed than comparison approaches.  

\begin{table}
\caption{Model complexity is evaluated on the Kodak dataset. Inference time is tested using a single NVIDIA GeForce RTX 4090 GPU. Best results are shown in \textbf{bold}.}
\setlength\tabcolsep{1pt}
\label{tab:1}
\centering
\begin{tabular}{lccccc}
\toprule
{Methods} & {Params (M)} & {FLOPs (G)} & {Enc (ms)} & {Dec (ms)} & {Avg (ms)} \\
\midrule
HiFiC     & 181.57 & 242.10 & 214.59 & 565.44   & 390.02  \\
MS-ILLM   & 181.48 & 293.06 & 43.11 & 53.09 & 48.10       \\
CDC       &  53.89 & 404.58 & 31.70 & 7,991.26  & 4,011.48 \\
TACO      & 101.75 & 232.50 & 131.21 & 68.58 & 99.90     \\
DiffEIC   & 1,379.50 & {--}   & 210.14 & 106.69 & 158.42 \\
RDEIC     & 1,380.27 & {--} & 153.06 & 93.68 & 123.37 \\
OSDiff    &  1,766.73 & {--} & 179.87 & 111.90 & 145.89 \\
ICISP     & 29.26 & 76.06 & 52.70 & 56.69 & 54.70 \\
\midrule
SAMIC (Ours) & \textbf{24.88} & \textbf{53.41} & \textbf{22.25} & \textbf{15.83} & \textbf{19.04} \\
\bottomrule
\end{tabular}
\end{table}

\subsection{Ablation study}

    

\begin{table}
\centering
\setlength\tabcolsep{1pt}
\caption{Contribution of each component for compression.}
\label{tab:2}
\begin{tabular}{lcccc}
\toprule
{Exp.} & {Methods} & {BD-Rate (\%) ↓} & {Params (M)} & {FLOPs (G)} \\
\midrule
E1 & SwinT + SVD-RRM          & 72.57 & 38.70 & 80.83 \\
E2 & SS2D + SVD-RRM           & 4.81 & 18.94 & 28.35     \\
E3 & SS2D                 & 9.12 & 18.94 & 28.35    \\
E4 & SASS                     & 1.22 & 24.88 & 53.41    \\ 
\midrule
E5 & SASS + SVD-RRM                 & 0  & 24.88 & 53.41 \\
\bottomrule
\end{tabular}
\end{table}

To systematically assess the contribution of each component in the proposed SAMIC method, 
we conducted ablation studies on the Kodak dataset. 

\subsubsection{Effect of Semantic-Aware Selective Scanning}
To validate the advantage of the proposed SASS, we replaced it with the original four-directional SS2D \cite{VisionMamba_ICML2024}. 
As shown in Table~\ref{tab:2} (E3 vs. E4), using the proposed SASS outperforms the original SS2D by achieving a bitrate savings of 7.9\%. This highlights that our scanning mechanism significantly improves compression performance by ensuring semantic continuity.

%
%
\subsubsection{Impact of SVD-RRM}
We analyzed the contribution of the SVD-RRM by removing it from the proposed method. Table~\ref{tab:2} shows that using SVD-RRM saves $1.22\%$ more bits and introduces negligible parameters and FLOPs (see E4 vs. E5). 
This confirms that SVD-RRM effectively removes channel redundancy via low-rank approximation, providing coding gains with nearly zero additional inference cost.

\subsubsection{Comparison with Other Nonlinear Transforms}
We further compare our method with Swin Transformer-based (SwinT for short) \cite{liu2021swin} nonlinear transforms which are widely used in existing compression methods \cite{lu2021transformer, MTCM_CVPR2023}.
Table~\ref{tab:2} (E1 vs. E5) demonstrates that the proposed SAMIC achieves bitrate savings of 72.57\% meanwhile has lower model complexity, further indicating that our method offers a far superior trade-off between compression performance and computational efficiency.

\subsubsection{Visualization of ERF} 
\begin{figure}
    \centering
    \begin{minipage}{0.26\linewidth}
        \includegraphics[width=\linewidth]{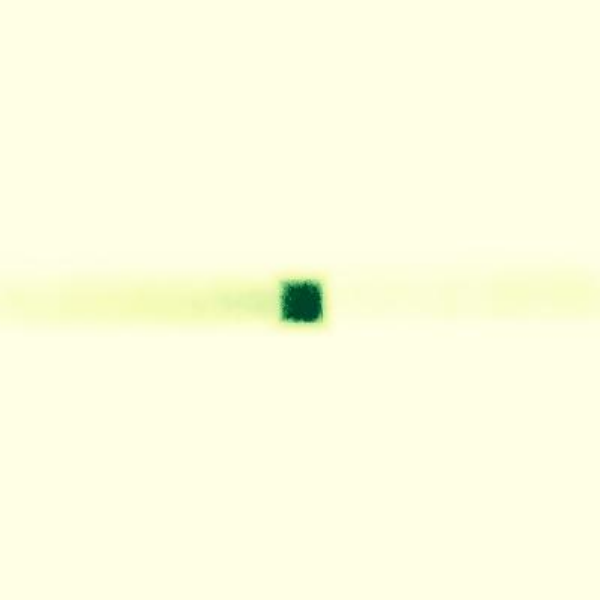}
                \centerline{SASS+SVD-RRM}
    \end{minipage}
    \hfill
    \begin{minipage}{0.26\linewidth}
        \includegraphics[width=\linewidth]{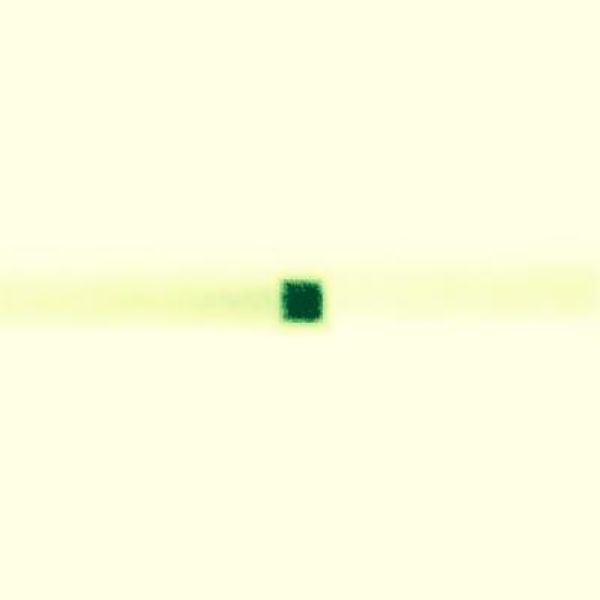}
                \centerline{SASS}
    \end{minipage}
    \hfill
    \begin{minipage}{0.26\linewidth}
        \includegraphics[width=\linewidth]{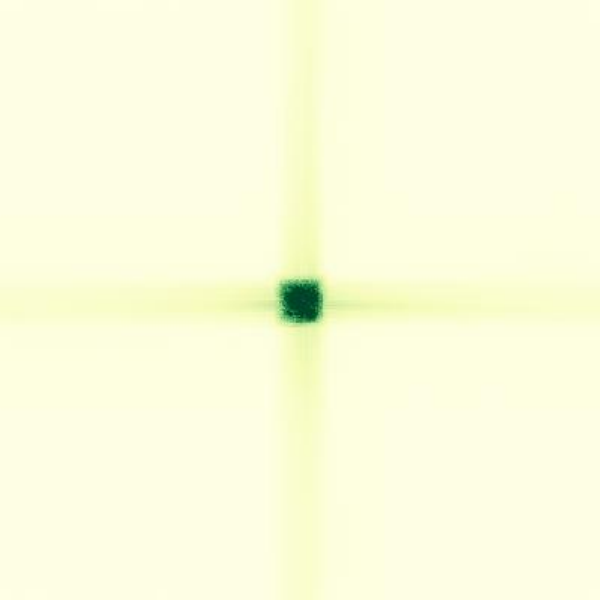}
                \centerline{SS2D}
    \end{minipage}
    \hfill
    \begin{minipage}{0.09\linewidth}
        \includegraphics[width=\linewidth, height=2.4cm]{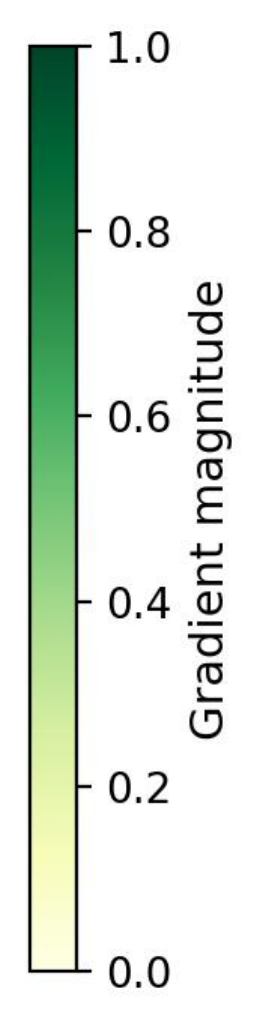}
                \centerline{}
    \end{minipage}
    \caption{Visualization of effective receptive field. Zoom in for a better view.}
    \label{fig:7}
\end{figure}

We further visualize the effective receptive field (ERF) in Fig.~\ref{fig:7}. Compared to the SS2D, our SASS exhibits a significantly more pronounced horizontal expansion. This broader horizontal ERF demonstrates that SASS can establish contextual dependencies across a wider range of pixels, which is instrumental for capturing high-frequency structural textures and eliminating spatial redundancies. Notably, introducing SVD-RRM does not compromise the model's inherent long-range modeling capabilities, effectively preserving the robustness of the SASS in spatial scanning (see "SASS+SVD-RRM" and "SASS").

\subsubsection{Analysis of Latent Correlation}
\begin{figure}
    \centering

    \includegraphics[width=0.30\linewidth,height=2.5cm]{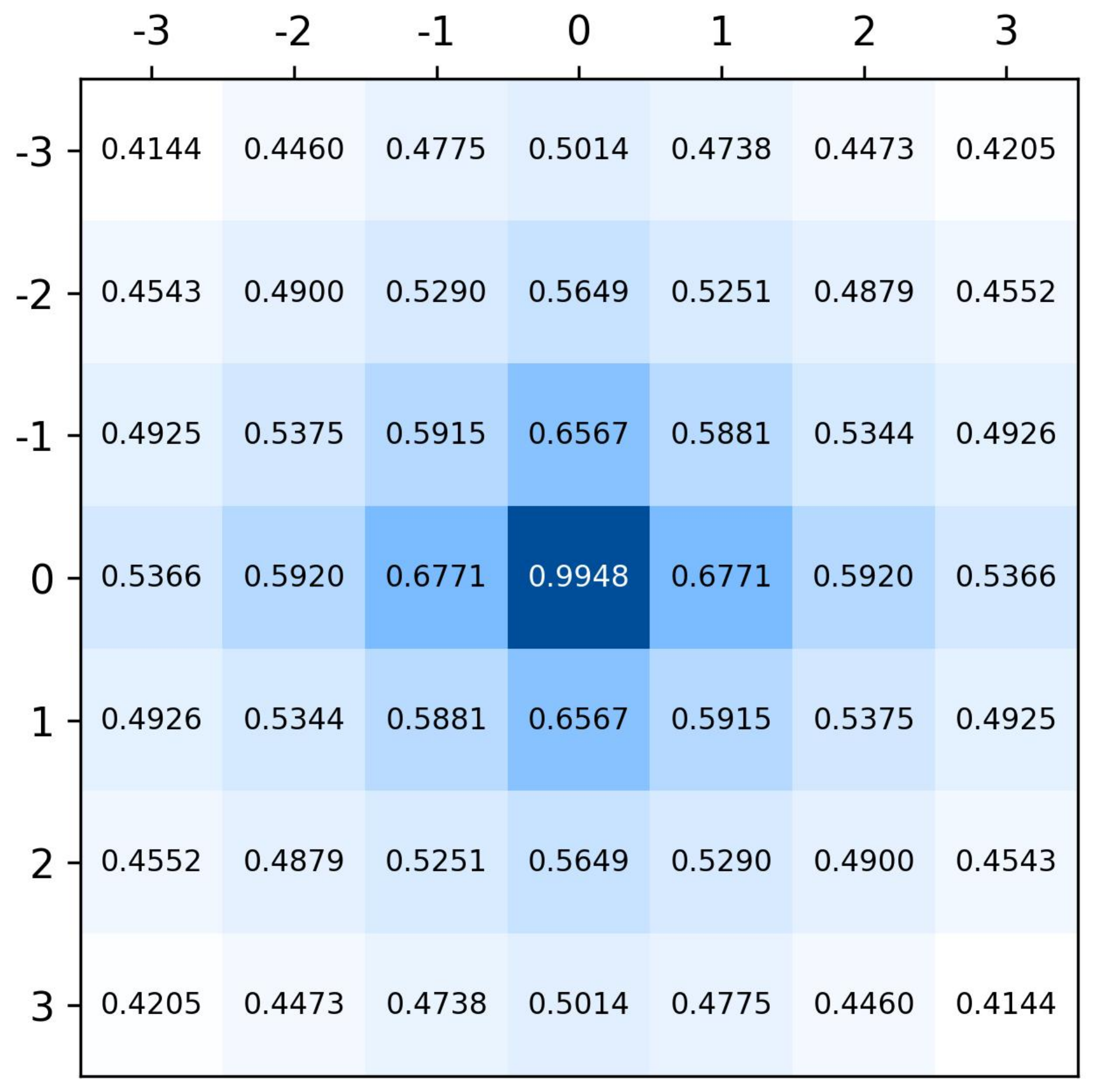}
    \includegraphics[width=0.30\linewidth,height=2.5cm]{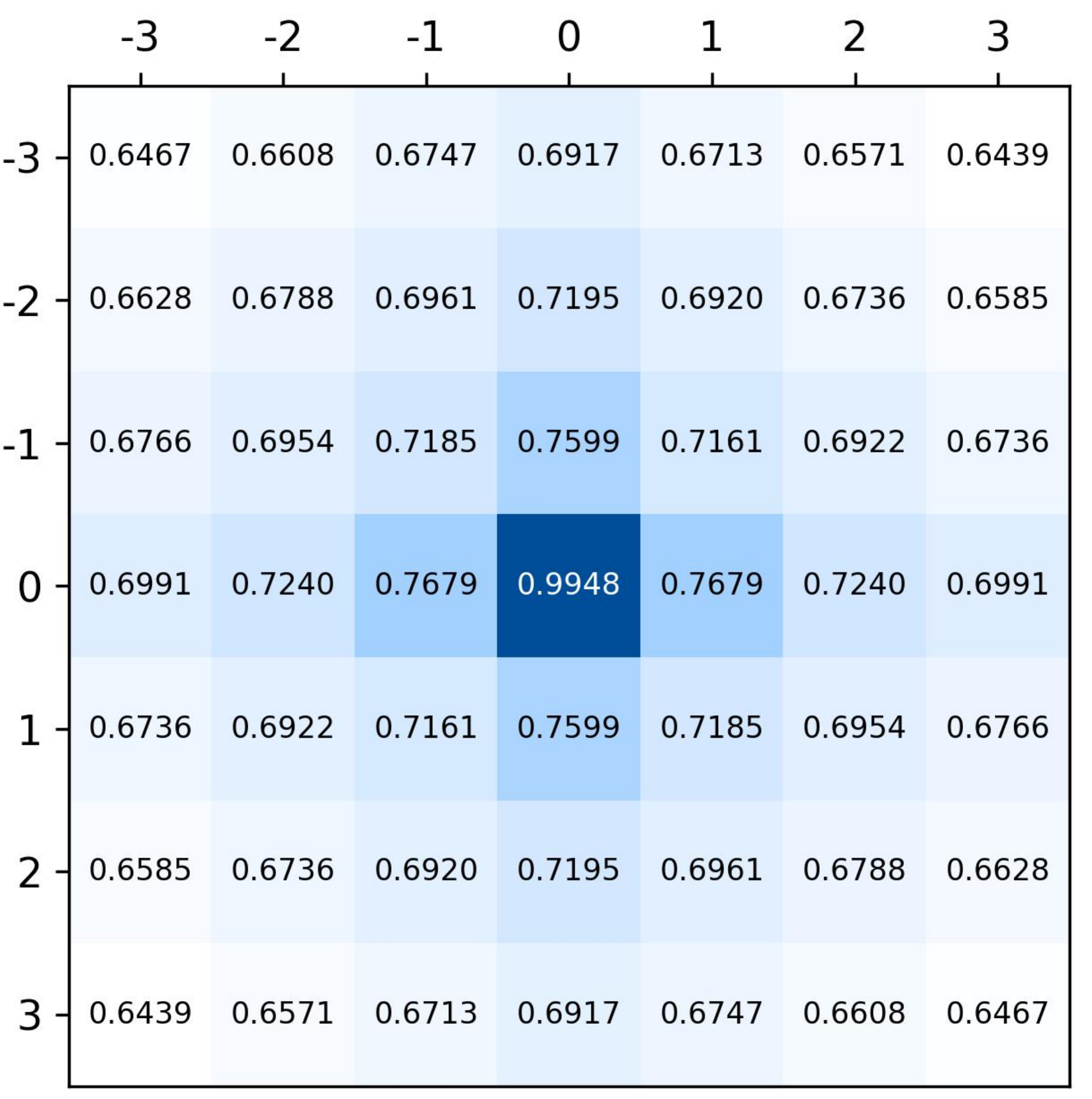}
    \includegraphics[width=0.34\linewidth,height=2.5cm]{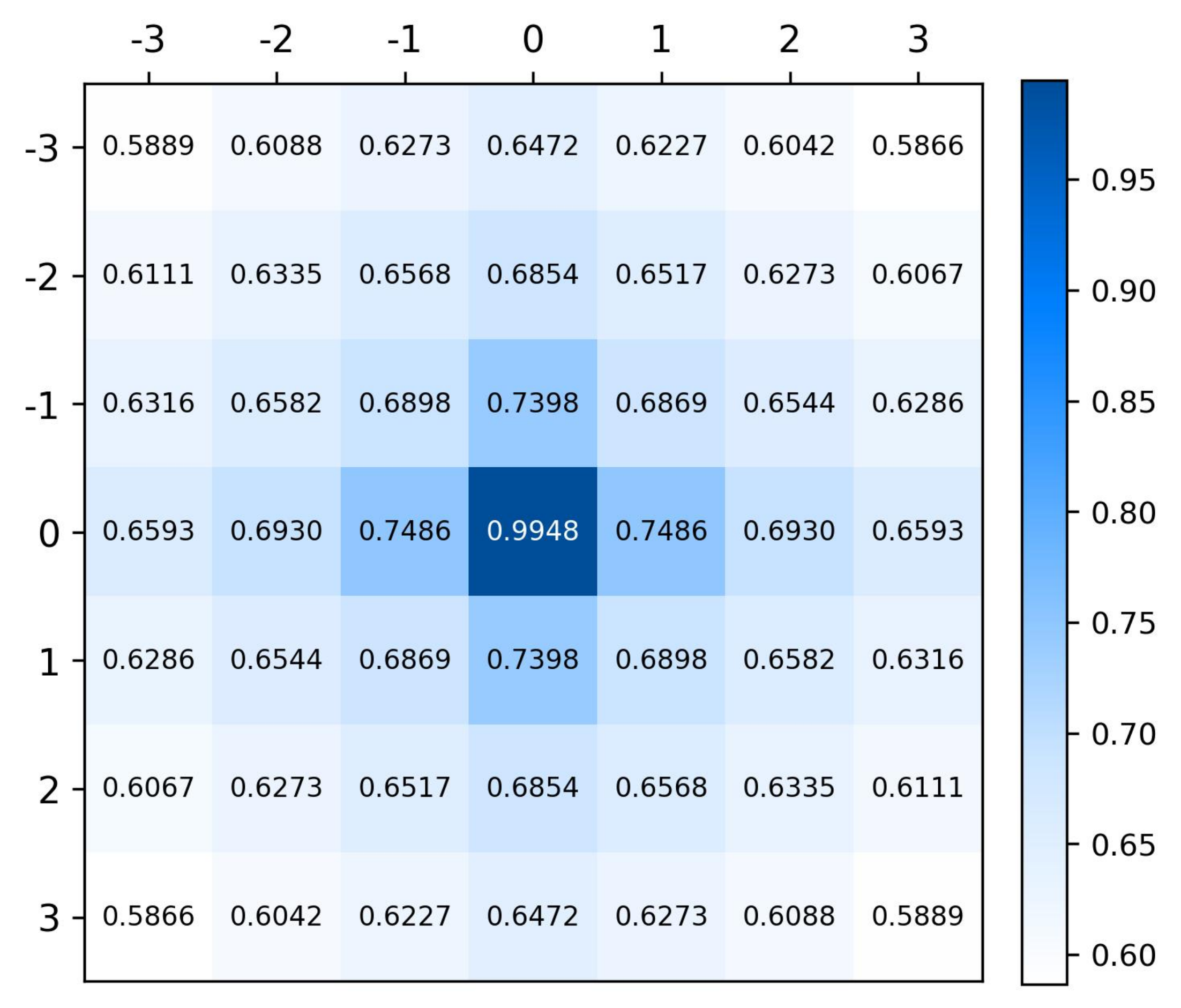}
    \makebox[0.32\linewidth]{SAMIC (Ours)}
    \makebox[0.32\linewidth]{SS2D}
    \makebox[0.32\linewidth]{Swin Transformer}
    \caption{Latent correlation of $(\text{y}-\mu)/\sigma$ with training $\lambda=0.025$.}
    \label{fig:8}
\end{figure}

To evaluate the efficacy of spatial redundancy removal, we visualize the latent correlation of the normalized representation $\tilde{\text{y}} \triangleq (\text{y}-\mu)/\sigma$. From an information-theoretic perspective, the primary objective of the nonlinear transform and conditional probability modeling is to decorrelate the latent features. As illustrated in  Fig.~\ref{fig:8}, our SAMIC consistently exhibits lower correlation coefficients across all spatial distances compared to the SS2D-based and Swin Transformer-based methods. 
This reduced correlation provides empirical evidence that our model eliminates spatial redundancies more effectively, leading to lower bitrate consumption. 
%
\subsubsection{Limitation}
\begin{figure}
    \centering
    \includegraphics[width=0.98\linewidth]{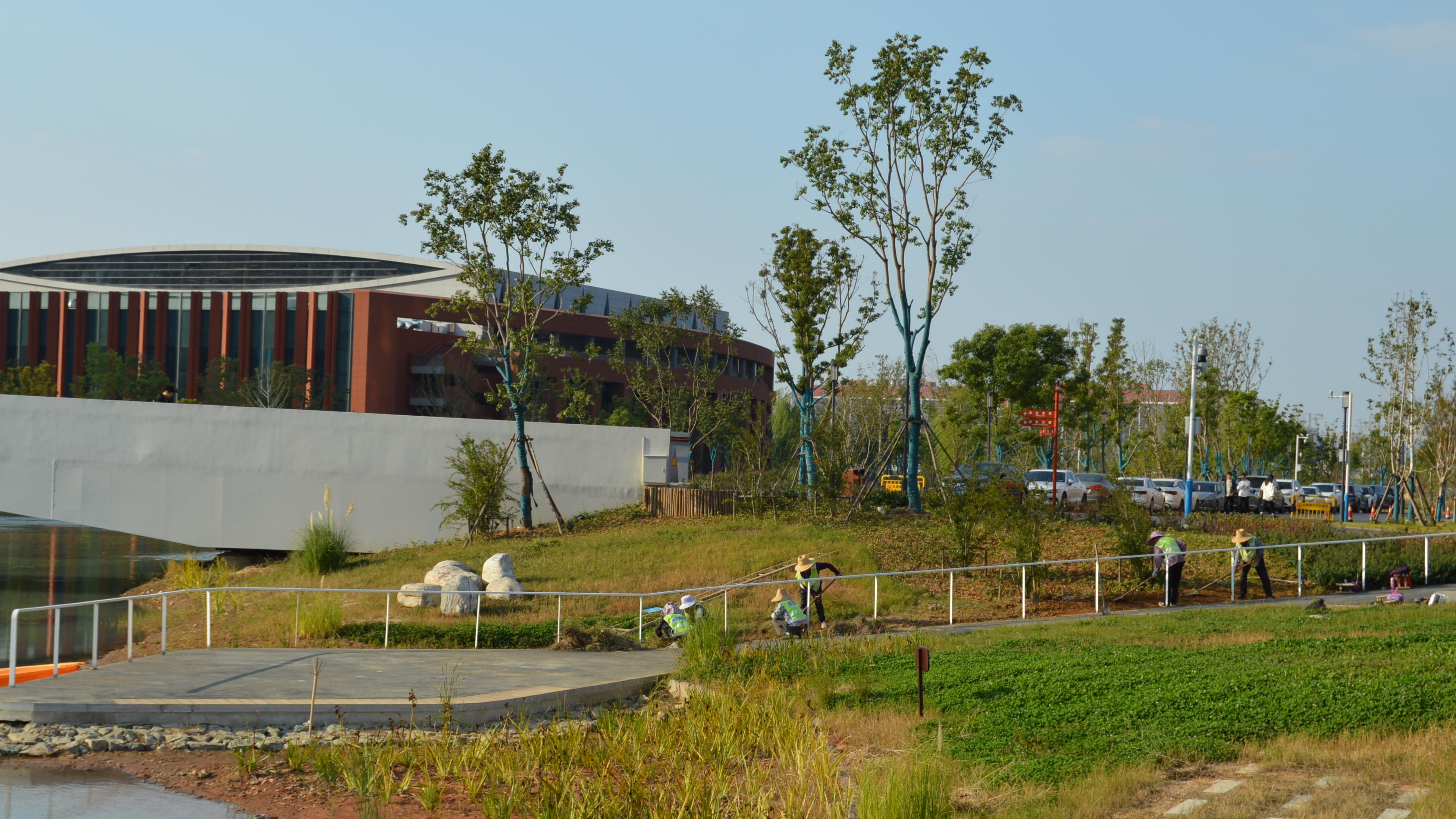} 
    \makebox[0.32\linewidth]{Original 4K-resolution Image} \\
    \includegraphics[width=0.32\linewidth]{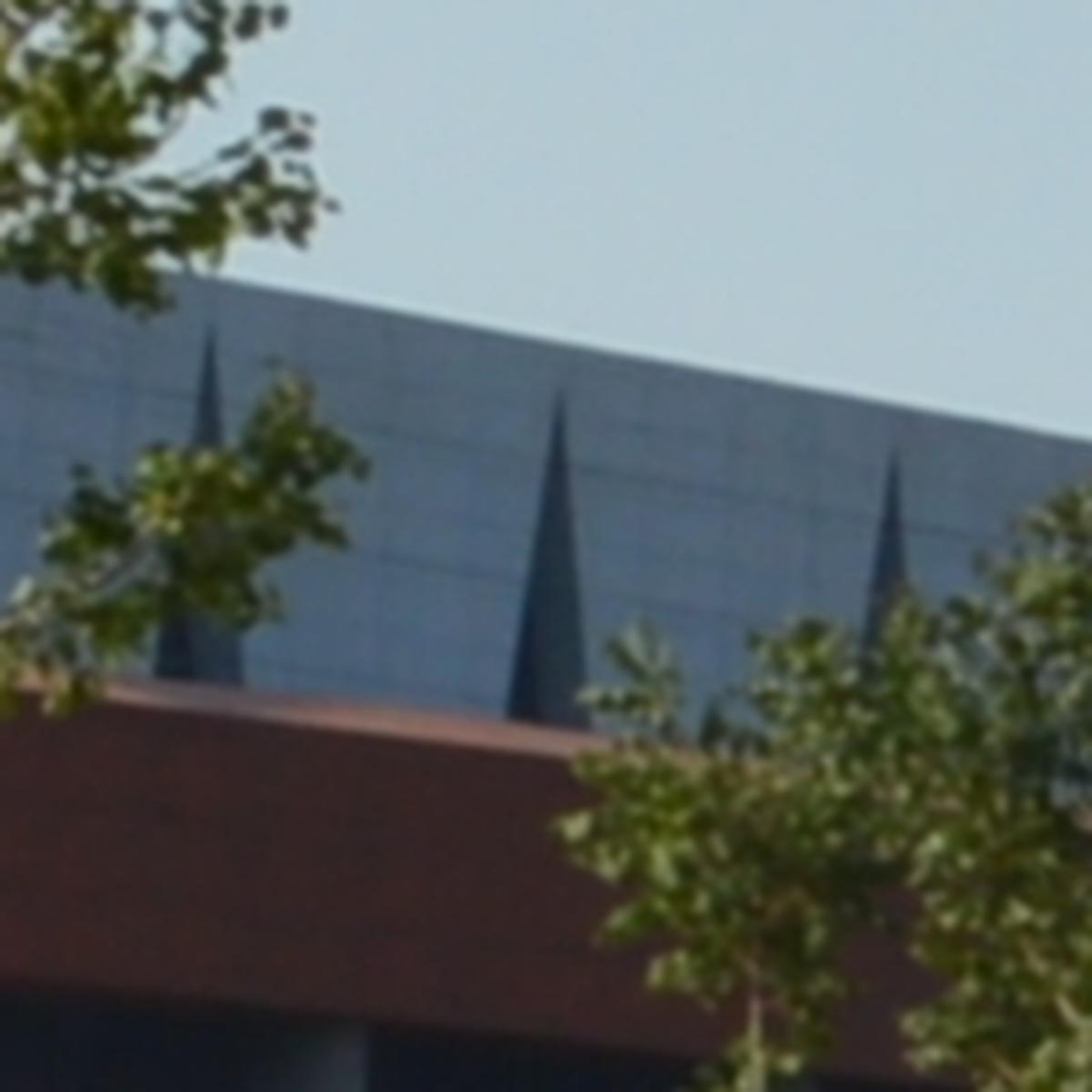}           
    \includegraphics[width=0.32\linewidth]{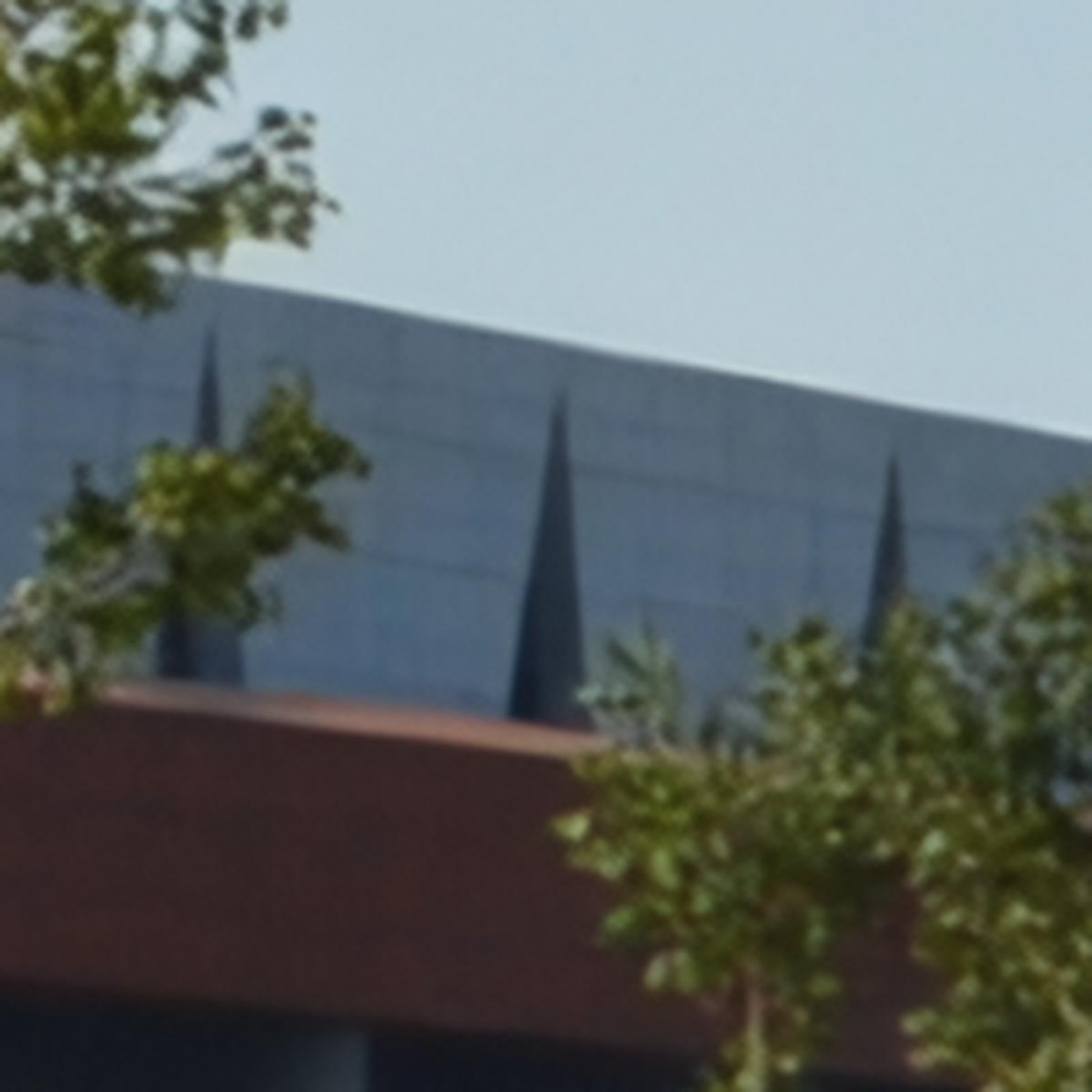}
    \includegraphics[width=0.32\linewidth]{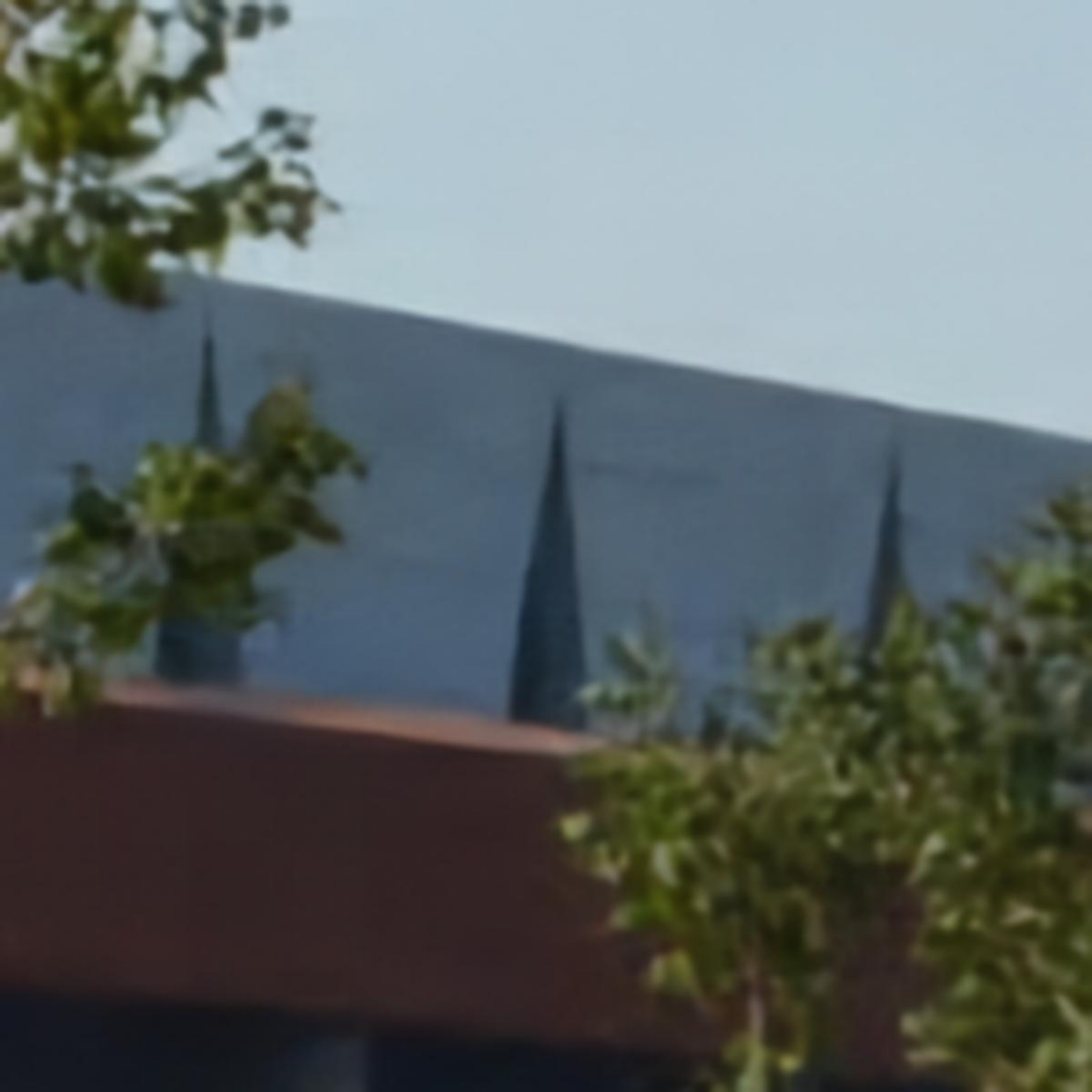} \\
    \makebox[0.32\linewidth]{GT}
    \makebox[0.32\linewidth]{ICISP (0.333)}
    \makebox[0.32\linewidth]{SAMIC (0.346)} \\
    \caption{Visual example when compessing a 4K-resolution image.}
    \label{fig:9}
\end{figure}

We further evaluate the proposed SAMIC method on the ultra-high-definition (UHD) dataset \cite{li2025ustc}. For a 4K-resolution image, our method requires 53.41 GFLOPs, with encoding and decoding times of 389.43 ms and 374.32 ms, respectively. Fig.~\ref{fig:9} shows a visual comparison where our method produces over-smoothed results, leading to loss of building structure. This issue arises because the UHD image contains diverse semantic categories, which prevent our method from forming accurate clusters before scanning. Future work will explore more fine‑grained grouping solutions to address this issue.

\section{Conclusion}
In this paper, we proposed SAMIC, a lightweight and efficient perceptual image compression framework based on the state space model. 
To address the inherent limitations of fixed scanning orders in traditional Mamba architectures, we developed the semantic-aware Mamba block (SAMB). By dynamically generating an adaptive scanning order guided by clustered semantic features, SAMB ensures both semantic and spatial continuity while effectively mitigating the strict causality constraints and long-range information decay inherent to Mamba. 
Furthermore, we designed an SVD-inspired redundancy reduction module (SVD-RRM) incorporated within the encoder. By introducing a learnable soft threshold for low-rank approximation, the SVD-RRM explicitly reduces channel redundancy with virtually no additional inference cost. 
Extensive experiments demonstrate that our approach achieves a superior rate-distortion-perception tradeoff compared to state-of-the-art methods, while enjoying the lowest model complexity.




\bibliographystyle{ACM-Reference-Format}
\bibliography{ref}

\newpage
\appendix
\begin{figure*}[t]
    \centering   
    \includegraphics[width=\textwidth]{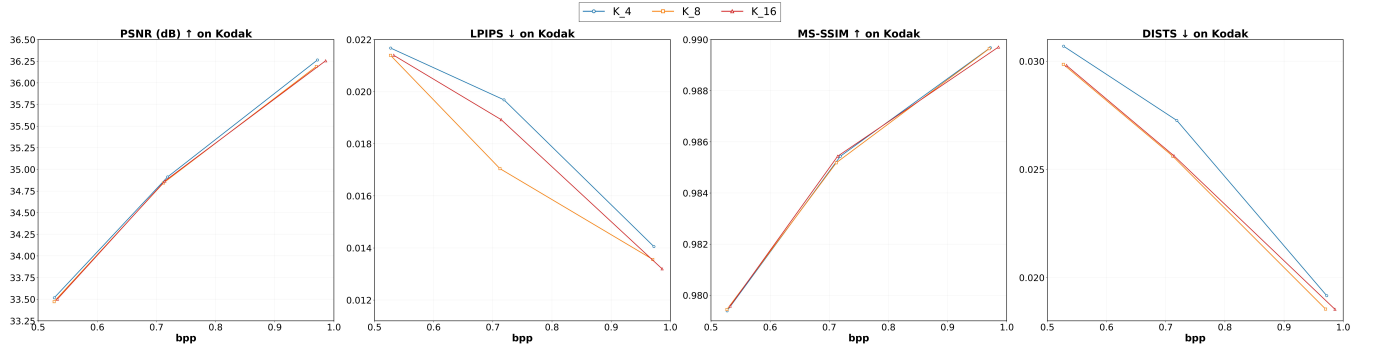}      
    \caption{Quantitative performance comparisons under different semantic cluster numbers ($K$) on the Kodak dataset.}
    \label{fig:10}
\end{figure*}

\begin{figure*}[t]
    \centering
    \begin{minipage}[b]{\linewidth}
        
        \begin{minipage}[b]{0.99\linewidth}
            \centering
            \begin{minipage}[b]{0.13\linewidth}
                \includegraphics[width=\linewidth]{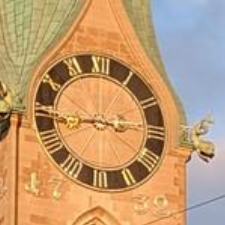}
                \centerline{(a3) GT (24)}
            \end{minipage}
            \hfill
            \begin{minipage}[b]{0.13\linewidth}
                \includegraphics[width=\linewidth]{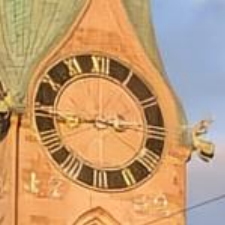}
                \centerline{(b3) SAMIC (0.155)}
            \end{minipage}
            \hfill
            \begin{minipage}[b]{0.13\linewidth}
                \includegraphics[width=\linewidth]{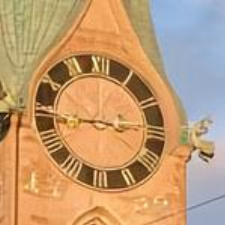}
                \centerline{(c3) TACO (0.178)}
            \end{minipage}
            \hfill
            \begin{minipage}[b]{0.13\linewidth}
                \includegraphics[width=\linewidth]{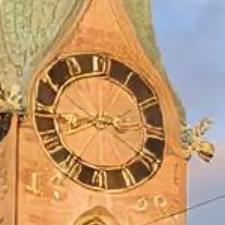}
                \centerline{(d3) CDC (0.227)} 
            \end{minipage}
            \hfill
            \begin{minipage}[b]{0.13\linewidth}
                \includegraphics[width=\linewidth]{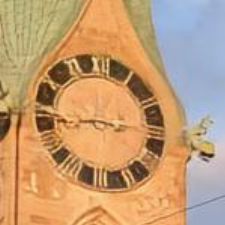}
                \centerline{(e3) HiFiC (0.216)}
            \end{minipage}
            \hfill
            \begin{minipage}[b]{0.13\linewidth}
                \includegraphics[width=\linewidth]{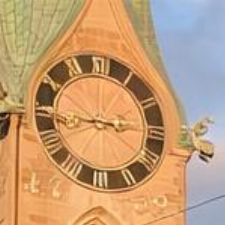}
                \centerline{(f3) ICISP (0.212)}
            \end{minipage}
            \hfill
            \begin{minipage}[b]{0.13\linewidth}
                \includegraphics[width=\linewidth]{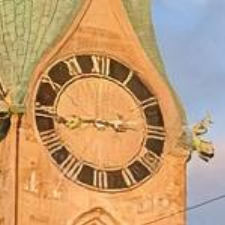}
                \centerline{(g3) MS-ILLM (0.202)}
            \end{minipage}
        \end{minipage}

        \vspace{4pt}

        \begin{minipage}[b]{0.99\linewidth}
            \centering
            \begin{minipage}[b]{0.13\linewidth}
                \includegraphics[width=\linewidth]{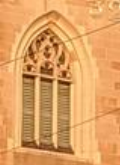}
                \centerline{(a4) GT (24)}
            \end{minipage}
            \hfill
            \begin{minipage}[b]{0.13\linewidth}
                \includegraphics[width=\linewidth]{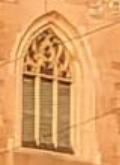}
                \centerline{(b4) SAMIC (0.155)}
            \end{minipage}
            \hfill
            \begin{minipage}[b]{0.13\linewidth}
                \includegraphics[width=\linewidth]{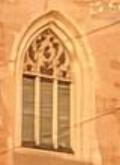}
                \centerline{(c4) TACO (0.178)}
            \end{minipage}
            \hfill
            \begin{minipage}[b]{0.13\linewidth}
                \includegraphics[width=\linewidth]{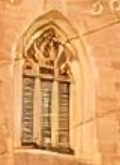}
                \centerline{(d4) CDC (0.227)} 
            \end{minipage}
            \hfill
            \begin{minipage}[b]{0.13\linewidth}
                \includegraphics[width=\linewidth]{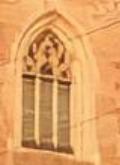}
                \centerline{(e4) HiFiC (0.216)}
            \end{minipage}
            \hfill
            \begin{minipage}[b]{0.13\linewidth}
                \includegraphics[width=\linewidth]{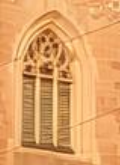}
                \centerline{(f4) ICISP (0.212)}
            \end{minipage}
            \hfill
            \begin{minipage}[b]{0.13\linewidth}
                \includegraphics[width=\linewidth]{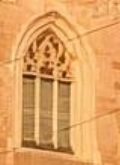}
                \centerline{(g4) MS-ILLM (0.202)}
            \end{minipage}
        \end{minipage}
    \end{minipage}
    \caption{\textbf{Visual comparisons on the CLIC2020 dataset.} The value inside the parentheses represents bpp.}
    \label{fig:11}
\end{figure*}

\section{Effect of the Number of Clusters}

In the main paper, we proposed a lightweight Semantic-Aware Selective Scanning (SASS) strategy, which dynamically groups image patches through soft clustering operations. The hyperparameter $K$ represents the number of learnable cluster centers, directly determining the semantic granularity to which the feature space is partitioned. To validate the rationality of the default configuration $K=16$ in the baseline network, this section conducts a comprehensive ablation study on $K \in \{4, 8, 16\}$ using the Kodak dataset, thoroughly evaluating its impact on the Rate-Distortion-Perception performance and actual computational overhead.
\subsection{Rate-Distortion-Perception Tradeoffs}
As shown in Fig.~\ref{fig:10}, across different bpp, the PSNR and MS-SSIM curves for the three configurations ($K=4$, $K=8$, and $K=16$) almost completely overlap. This indicates that the Mamba architecture itself possesses robust global modeling capabilities, sufficient to macroscopically minimize pixel-level mean square errors. Basic objective distortion metrics are largely insensitive to the semantic partitioning granularity of the local scanning trajectory. In the visual perception evaluation system, the selection of the $K$ value plays a decisive role in reconstruction quality. When coarse-grained clustering $K=4$ is adopted, the LPIPS and DISTS errors are significantly higher, showing obvious performance degradation, especially in the low-bitrate range. Due to the insufficient number of clusters, image semantics with vastly different visual characteristics are forced into the same cluster. This leads to a massive amount of semantic abruptions during the 1D sequential rearrangement, severely disrupting the continuity assumption during Mamba scanning. When the number of clusters is increased to 8 or 16, the perceptual metrics achieve a drastic improvement. Finer-grained cluster centers effectively separate feature flows of different attributes, making the intra-sequence contextual correlation purer. On the Kodak test set, the perceptual performance of $K=8$ and $K=16$ is very close, with both effectively preserving high-frequency structures.
\subsection{Computational Overhead Analysis}
To comprehensively evaluate the engineering deployment feasibility of the algorithm, we summarized the model parameters, FLOPs, and actual encoding/decoding latency in Table~\ref{tab:3}. As the value of $K$ doubles from 4 to 16, the total model parameters only marginally increase from 24.86 M to 24.88 M, and the FLOPs remain stable at around 53.40 G. This strongly proves that the proposed SASS module is extremely lightweight in architectural design, increasing the number of cluster centers do not incur additional parameter inflation or theoretical computational burden. In conventional understanding, an increase in the number of clusters is usually accompanied by a rise in computation time. However, the actual hardware inference results exhibit non-linear characteristics. On the RTX 4090 GPU, $K=16$ demonstrates an overwhelming speed advantage, with its average encoding/decoding time being only 19.04 ms, achieving significant acceleration compared to $K=8$ and $K=4$. 

Synthesizing the experimental data above, although $K=8$ can provide comparable perceptual quality on medium-to-low resolution test sets like Kodak, $K=16$ achieves an absolute advantage in actual inference efficiency. Moreover, considering the future need for the model to generalize to high-resolution or Ultra-High-Definition (UHD) images with complex semantic targets, $K=16$ provides a more abundant feature representation capacity, effectively preventing misclassification and texture over-smoothing phenomena in complex scenes. Therefore, we establish $K=16$ as the optimal hyperparameter configuration for this framework.

\begin{table}
\caption{Quantitative comparison of computational complexity and inference latency under different semantic cluster numbers $K$. Inference time is tested using a single NVIDIA GeForce RTX 4090 GPU. }
\setlength\tabcolsep{1pt}
\label{tab:3}
\centering
\begin{tabular}{lccccc}
\toprule
{K} & {Params (M)} & {FLOPs (G)} & {Enc (ms)} & {Dec (ms)} & {Avg (ms)} \\
\midrule
4 & 24.86 & 53.40 & 43.10 & 23.85 & 33.48 \\
8 & 24.87 & 53.40 & 65.27 & 32.81 & 49.04 \\
16 & 24.88 & 53.41 & 22.25 & 15.83 & 19.04 \\
\bottomrule
\end{tabular}
\end{table}

\section{More Visual Examples}

To further validate the generalization capability of the SAMIC method, we conducted qualitative comparisons on additional test samples. As shown in Fig.~\ref{fig:11}, visual comparisons on the CLIC2020 dataset demonstrate that SAMIC maintains clear texture details of clock faces and architectural structures even at low bitrates (0.155 bpp), whereas competing methods exhibit varying degrees of blurring or artifacts at comparable or higher bitrates. Particularly in high-frequency detail regions, such as Roman numerals on clock faces and window grilles, SAMIC demonstrates superior fidelity. Meanwhile, we can also observe the limitation of our method, in high-frequency detail regions such as window grilles, SAMIC produces smoother results compared to ICISP.

\end{document}